%% file: main.tex
\pdfoutput=1
\documentclass[11pt]{article}
\usepackage[preprint]{acl}
\newif\ifpreprint\preprinttrue
\input{preamble}
\begin{document}
\input{body}

\end{document}

%% file: preamble.tex
\usepackage{times}
\usepackage{latexsym}
\usepackage[T1]{fontenc}
\usepackage[utf8]{inputenc}
\usepackage{microtype}
\usepackage{inconsolata}
\usepackage{graphicx}
\usepackage{booktabs}
\usepackage{multirow}
\usepackage{amsmath,amssymb,amsthm}
\usepackage{xcolor}
\usepackage{enumitem}
\usepackage{tikz}
\usepackage{stfloats}
\usetikzlibrary{positioning,arrows.meta}

\newsavebox{\fittedtable}
\newcommand{\fittable}[1]{%
  \sbox{\fittedtable}{#1}%
  \ifdim\wd\fittedtable>\linewidth
    \resizebox{\linewidth}{!}{\usebox{\fittedtable}}%
  \else
    \usebox{\fittedtable}%
  \fi
}

\input{macros}
\allowdisplaybreaks[1]

\input{tables/generated_numbers}

\title{From Retrieval to Typed Decisions:\\
Calibrated System One Models from Biomedical Sentence Encoders}

\author{Pritam Deka \\
  Queen's University Belfast \\
  Belfast, United Kingdom}

%% file: macros.tex
\newcommand{\ours}{\textsc{sbert2s1}}
\newcommand{\bench}{\textsc{BioDecide}}
\newcommand{\medline}{\textsc{MEDLINE-S1}}

\newcommand{\pfr}{PFR}
\newcommand{\choice}{\texttt{choice}}
\newcommand{\score}{\texttt{score}}
\newcommand{\noul}{\texttt{noul}}

\newtheorem{proposition}{Proposition}
\DeclareMathOperator*{\argmax}{arg\,max}
\newcommand{\E}{\mathbb{E}}
\newcommand{\ind}{\mathbb{I}}
\newcommand{\softmax}{\operatorname{softmax}}

%% file: tables/generated_numbers.tex
\newcommand{\CascadeSCE}{68.4}
\newcommand{\AloneSCE}{59.5}
\newcommand{\CascadeL}{69.1}
\newcommand{\AloneL}{64.3}
\newcommand{\AloneGemma}{62.1}

%% file: body.tex
\maketitle
\input{sections/abstract}
\input{sections/introduction}
\input{sections/related}
\input{sections/method}
\input{sections/data}
\input{sections/setup}
\input{sections/results}
\input{sections/discussion}

\input{sections/acknowledgements}
\input{sections/declarations}

\bibliography{references}

\appendix
\section*{Appendix}
\renewcommand{\topfraction}{0.95}
\renewcommand{\dbltopfraction}{0.95}
\renewcommand{\textfraction}{0.05}
\renewcommand{\floatpagefraction}{0.65}
\renewcommand{\dblfloatpagefraction}{0.65}
\setcounter{topnumber}{3}
\setcounter{dbltopnumber}{3}
\input{sections/appendix}

%% file: sections/abstract.tex
\begin{abstract}
Typed decision models answer schema-constrained questions about a text in one forward pass and
return probabilities meant to be thresholded. We ask whether biomedical sentence encoders
trained for retrieval are good starting points for such models. We present \ours{}, which converts
Sentence-Transformers encoders into bi-encoder, cross-head (C) and prior-fused residual (\pfr{})
decision models, together with \bench{}, a biomedical typed-decision suite, and \medline{}, 243k
training decisions derived from NLM indexing. Across six parent-retriever pairs, retrieval training
improves zero-shot matching of content-bearing options. After fine-tuning, its effect depends on the
head: across five pairs and three training-set sizes, retrieval training significantly helps \pfr{},
which keeps the retrieval prior, in 10 of 15 comparisons, but helps C in one and hurts it in five. A matched grid of two heads and
five training objectives shows that C outperforms \pfr{} under every objective, and that the released
policy-gradient recipe trails cross-entropy by 2.5--3.0 points. The deficit appears
to stem mainly from its reward normalisation, which inflates the noisy score-function term 3.6--15-fold; an
unbiased leave-one-out estimator recovers most of the gap. After temperature scaling, no objective
is clearly better calibrated than cross-entropy. In a low-label ablation on one encoder, a fixed
cosine prior gave no consistent accuracy advantage over shuffled scores and underperformed the
residual alone. Refitting temperatures after document filtering yielded a small loss in
seen-task log likelihood and a small improvement on held-out tasks. We release the code,
the \medline{} labels and a model.
\end{abstract}

%% file: sections/introduction.tex
\section{Introduction}\label{sec:intro}
Much of the language processing in biomedical software consists of small, bounded decisions: does this
abstract report a randomised trial, does the evidence support the claim, how severe are the side effects
a patient describes? Because the answers are consumed by code, a decision is only useful if it respects
a schema and comes with a confidence that can be thresholded to act, defer or escalate.

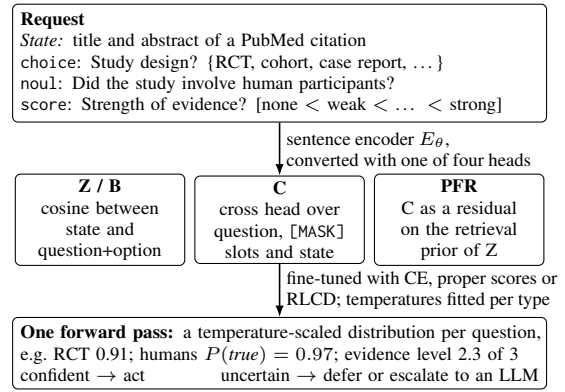
\begin{figure}[t]
\centering\scriptsize
\begin{tikzpicture}[
  box/.style={draw, rounded corners=2pt, align=left, inner sep=3pt, text width=6.85cm},
  conv/.style={draw, rounded corners=2pt, align=center, inner sep=2pt, text width=2.1cm, minimum height=1.05cm},
  arr/.style={-{Latex[length=1.5mm]}, semithick}]
\node[box] (req) {\textbf{Request}\\
  \emph{State:} title and abstract of a PubMed citation\\
  \texttt{choice}: Study design? \{RCT, cohort, case report, \ldots\}\\
  \texttt{noul}: Did the study involve human participants?\\
  \texttt{score}: Strength of evidence? [none $<$ weak $<$ \ldots{} $<$ strong]};
\node[conv, below=7mm of req.south, anchor=north] (c) {\textbf{C}\\ cross head over\\ question, \texttt{[MASK]}\\ slots and state};
\node[conv, left=1.2mm of c] (zb) {\textbf{Z / B}\\ cosine between\\ state and\\ question +\\ option};
\node[conv, right=1.2mm of c] (pfr) {\textbf{\pfr{}}\\ C as a residual\\ on the retrieval\\ prior of Z};
\node[box, below=7mm of c.south, anchor=north] (out) {\textbf{One forward pass:} a temperature-scaled distribution per question,
  e.g.\ RCT 0.91; humans $P(\textit{true})=0.97$; evidence level 2.3 of 3\\
  confident $\rightarrow$ act \hfill uncertain $\rightarrow$ defer or escalate to an LLM};
\draw[arr] (req.south) -- node[right, align=left, font=\scriptsize] {sentence encoder $E_\theta$,\\ converted with one of four heads} (c.north);
\draw[arr] (c.south) -- node[right, align=left, font=\scriptsize] {fine-tuned with CE, proper scores or\\ RLCD; temperatures fitted per type} (out.north);
\end{tikzpicture}
\caption{Overview. A System One request pairs a state with typed questions. A biomedical sentence encoder,
converted with one of four heads (\S\ref{sec:method}), returns a calibrated distribution over each question's
options in one forward pass, and downstream code thresholds it to act or escalate. Example values are
illustrative.}
\label{fig:overview}
\end{figure}
\emph{System One} models were proposed for exactly this interface~\citep{typesafe2026jev}. A request
pairs a text (the \emph{state}) with typed questions of three kinds (\choice{}, \score{} and \noul{}),
and the model returns, in one forward pass, a distribution over the options each question allows. The open model
Laya~\citep{laya2026} pairs an encoder with a mask-slot head and trains it with RLCD, a mixture of
cross-entropy and a policy gradient against proper scoring rules~\citep{gneiting2007strictly}. To
our knowledge, all existing System One models are general-domain and start from masked-language-model (MLM) or decoder
checkpoints. Most biomedical encoders are MLMs too, such as BioBERT~\citep{lee2020biobert} and
PubMedBERT~\citep{gu2021domain}. A smaller group has been further trained contrastively as
Sentence-Transformers encoders for retrieval: S-PubMedBERT-MS-MARCO~\citep{deka2022improved} is
PubMedBERT fine-tuned on MS MARCO~\citep{nguyen2016ms}, and MedCPT~\citep{jin2023medcpt} is trained on PubMed search logs.
A typed decision resembles retrieval, since it asks whether a hypothesis formed from a question and an
option matches the state, but retrieval similarity need not encode negation, ordinal judgement or
calibrated truth. This study tests whether biomedical sentence encoders can be converted into System One
models, and what the result depends on:
\begin{enumerate}[label=\textbf{RQ\arabic*},leftmargin=*,itemsep=1pt,topsep=3pt]
\item \textbf{Initialisation:} when does retrieval training improve zero-shot or fine-tuned decisions?
\item \textbf{Conversion:} how do bi-encoder, cross-head and prior-fused conversions trade accuracy,
robustness to option order, cost and retrieval retention, under matched objectives?
\item \textbf{Objective:} which ingredient of RLCD matters once temperatures are fitted: the proper
score, the noise smoothing, or the policy-gradient estimator?
\end{enumerate}

\paragraph{Contributions.}
(1)~\ours{}, a framework that turns Sentence-Transformers encoders into typed decision models with four
conversions, including a \emph{prior-fused residual} (\pfr{}) head that starts from a scaled
cosine retrieval prior; (2)~\bench{}, a biomedical typed-decision suite with public and credentialed
clinical tracks, and \medline{}, multi-question training data labelled from NLM indexing rather than an
LLM teacher; (3)~a controlled study over eleven encoders, six parent-retriever pairs, and a matched
grid of two heads and five objectives, with follow-up tests of score assignment and calibration
under document filtering; and (4)~an analysis showing that RLCD optimises a
noise-smoothed proper score, that the released estimator is biased and rescaled, and that a simple
leave-one-out baseline removes both problems.
\ifpreprint
Code, \medline{} and the released model are available at
\url{https://github.com/pritamdeka/sbert2s1},
\url{https://huggingface.co/datasets/pritamdeka/MEDLINE-S1} and
\url{https://huggingface.co/pritamdeka/S1-PubMedBERT}.
\fi

%% file: sections/related.tex
\section{Related Work}
\label{sec:related}

\paragraph{System One models.}
Jev~\citep{typesafe2026jev} introduced typed decisions with three primitives: \choice{} (one option
from a schema), \score{} (a distribution over an ordinal rubric) and \noul{} (the probability that a
statement holds); its training objective is not public. Laya~\citep{laya2026}, the first open
implementation, pairs ModernBERT-large~\citep{warner2024modernbert} with a two-layer head that scores one
\texttt{[MASK]} slot per option, trained with cross-entropy plus a group-baselined policy gradient (as in
GRPO; \citealp{shao2024deepseekmath}) on noise-perturbed logits rewarded by proper scores.
Decider~\citep{decider2026} fine-tunes decoders, and the Decision Index~\citep{decisionindex2026}
benchmarks typed-decision engines. Concurrent work applies or probes System One models in security
agents, annotation, crash-report coding, distracting contexts and active
questioning~\citep{pentest2026,css2026,crash2026,jevout2026,lavoir2026} (Appendix~\ref{app:related}).
Two further preprints, posted in October 2026, study decision models for retrieval. \citet{dinzinger2026notbetter}
compare decision fine-tuning with ordinary reranker fine-tuning on Qwen backbones over TREC DL and BEIR
collections and report that decision fine-tuning does not give the better reranker, while
\citet{wan2026retrieverfree} use decision models to navigate a document hierarchy instead of a first-stage retriever.
Both build on language-model backbones rather than sentence encoders, and we do not compare against them.
To our knowledge, none of these studies evaluates biomedical conversions of sentence encoders, retrieval initialisation, or RLCD ablations.

\paragraph{Sentence encoders as decision models.}
Siamese encoders trained so that cosine similarity reflects relatedness~\citep{reimers2019sentence},
on query-passage data such as MS~MARCO~\citep{nguyen2016ms}, are standard for dense
retrieval~\citep{karpukhin2020dense}. Late interaction keeps token-level matching at a retrieval-friendly
cost~\citep{khattab2020colbert}, and cross-encoders re-rank more accurately at higher
cost~\citep{nogueira2019passage}. Our conversions sit along this trade-off: the bi-encoder B scores with a
precomputed state embedding, the cross head C reads question and state jointly, and the prior-fused residual
\pfr{} adds a cross score to the retrieval prior. Comparing a text with label descriptions underlies
entailment-based zero-shot classification~\citep{yin2019benchmarking}. In biomedicine, domain
pre-training~\citep{gu2021domain} followed by contrastive training on search
logs~\citep{jin2023medcpt} or MS~MARCO~\citep{deka2022improved} yields strong retrievers. We use such
encoders as the \emph{initialisation} of decision models and measure what the contrastive stage adds.

\paragraph{Cloze-style and unified prediction.}
Scoring an option at a \texttt{[MASK]} position relates System One heads to cloze-style prompting for
classification and inference~\citep{schick2021exploiting,gao2021making}, and answering questions in several
formats with one model has precedent in UnifiedQA~\citep{khashabi2020unifiedqa}. System One models instead return a
probability distribution over the options of a fixed schema in a single forward pass, without generating text,
which is the property that makes their confidence usable as a threshold.

\paragraph{Calibration and scoring rules.}
A strictly proper scoring rule is uniquely optimised in expectation by the true
distribution~\citep{gneiting2007strictly}; examples include the log, Brier~\citep{brier1950},
spherical and ranked-probability~\citep{epstein1969} scores. Post-hoc calibration of classifier
probabilities predates deep learning~\citep{niculescu2005predicting}. Neural networks, including pre-trained
transformers, are often miscalibrated, and temperature scaling is a strong post-hoc
remedy~\citep{guo2017calibration,desai2020calibration}. A calibration benefit claimed for a training
objective must therefore survive temperature scaling.

\paragraph{Policy-gradient estimators.}
Score-function (REINFORCE) estimators~\citep{williams1992simple} are unbiased but noisy, and a baseline reduces
their variance. Leave-one-out baselines, which subtract the mean reward of the other samples for the same
input, are used for REINFORCE-style training of language models~\citep{ahmadian2024back}. The PG-LOO objective
in this paper is therefore not a new estimator; our contribution is to show how the released RLCD recipe departs
from the unbiased version, by how much, and what that costs in accuracy.

%% file: sections/method.tex
\section{Method}
\label{sec:method}

\subsection{Typed decisions}
A request consists of a state $s$ (text or a serialised record) and questions $q=(\tau,x,o_{1:K})$,
where $\tau\in\{\choice,\score,\noul\}$ is the type, $x$ the instruction and $o_1,\dots,o_K$ the options:
schema keys with optional descriptions for \choice{}, ordered rubric levels for \score{}, and
$(\textit{false},\textit{true})$ for \noul{}. We use the request schema shared by Jev and Laya. A model
maps $(s,q)$ to logits $z\in\mathbb{R}^K$ and reports
\begin{equation}
  p(\cdot\mid s,q)=\softmax\big(z/T_{\tau,K}\big),
  \label{eq:report}
\end{equation}
where $T_{\tau,K}>0$ is a post-hoc temperature shared by all questions of type $\tau$ in one
option-count bucket (\S\ref{sec:calib}). The answer is $\argmax_k p_k$ for \choice{}, the expected level
$\sum_{k}(k-1)\,p_k$ for \score{}, and $p_2$ (\emph{true}) for \noul{}. The confidence $\max_k p_k$ is the
quantity that is thresholded and that expected calibration error (ECE) measures.

\subsection{Converting an encoder}
Let $E_\theta$ be the transformer of a Sentence-Transformers model and
\begin{equation}
  e_\theta(\cdot)=\frac{\operatorname{pool}(E_\theta(\cdot))}{\lVert\operatorname{pool}(E_\theta(\cdot))\rVert_2}
  \label{eq:embed}
\end{equation}
its normalised embedding, with the checkpoint's pooling (mean or \texttt{[CLS]}). For option $k$ we form
the \emph{hypothesis} $h_k=x\oplus o_k$ (instruction followed by the rendered option) and define the
retrieval prior $c_k=\langle e_\theta(s),e_\theta(h_k)\rangle$; the hypothesis plays the query and the
state the passage. We compare four conversions:
\begin{align}
  \text{Z, B:}\quad & z_k=\alpha_\tau\,c_k, \label{eq:bi}\\
  \text{C:}\quad & z_k=r_{\theta,\phi}(s,q)_k, \label{eq:cross}\\
  \text{\pfr{}:}\quad & z_k=r_{\theta,\phi}(s,q)_k+\alpha_\tau\,c_k. \label{eq:pfr}
\end{align}
\textbf{Z} keeps $\theta$ frozen and fits only the per-type scale $\alpha_\tau$. \textbf{B} trains
$\theta$ and $\alpha$; state embeddings can be pre-computed, so screening a corpus costs one embedding
of $K$ hypotheses and a matrix product. \textbf{C} is a cross-encoder head: $E_\theta$ encodes
\begin{align*}
  &\texttt{[CLS]}\,\tau\,x\,\texttt{[SEP]}\;\texttt{[MASK]}\,o_1\cdots\texttt{[MASK]}\,o_K\\
  &\qquad\texttt{[SEP]}\;s\,\texttt{[SEP]},
\end{align*}
a two-layer transformer $\phi$ with a type embedding contextualises it, and an MLP scores the hidden
state at the $k$-th \texttt{[MASK]}. The head replicates Laya's, so the two load each other's weights:
the public Laya checkpoint run through our implementation reproduces its reported zero-shot accuracy
on typed-decisions (0.3615 vs.\ 0.362). \textbf{\pfr{}} adds this cross score to the retrieval prior.
The last layer of the scorer is zero-initialised, so the untrained \pfr{} model equals Z and training
learns a cross-attention residual on top of the retriever. The prior shares $\theta$ by default; a
frozen copy preserves the original retriever at twice the memory. The prior scale is initialised at
$\alpha_\tau=20$, the usual similarity scale of contrastive encoders,\footnote{The implementation sets
$\alpha_\tau=\max(1/T^0_\tau,20)$ from a temperature $T^0_\tau$ fitted to raw cosines; because $T^0_\tau$ is
clipped to $[0.05,20]$, this is always 20.} has its own learning rate and no weight decay.

\subsection{Training objectives and the RLCD estimator}
\label{sec:rlcd}
All objectives compare the reported distribution $p$ with a target distribution $t$ (one-hot for most
items). The simplest is cross-entropy (CE). Laya's reward $S(p,t)$ adds two further proper scoring
rules to the log score: the spherical score for every question and, for \score{} questions, the ranked
probability score, which also penalises probability placed on distant rubric levels. A scoring rule is
\emph{proper} when reporting the true distribution maximises its expected value, so a high $S$ rewards
honest probabilities~\citep{gneiting2007strictly}.

RLCD does not maximise $S$ directly. For each item it adds Gaussian noise to the logits $G=4$ times,
scores each noisy prediction with $S$, and moves the logits towards the noise directions that scored
above average. This is a policy-gradient (REINFORCE) step, and CE is added to it with weight one. The
reward part of the update for one item is
\begin{equation}
  \hat g=\frac{1}{G\sigma^2}\sum_{g=1}^{G}\frac{R^{(g)}-\bar R}{d}\,\varepsilon^{(g)},
  \label{eq:pg}
\end{equation}
where $\varepsilon^{(g)}$ is the $g$-th noise draw, $\sigma$ its scale (annealed from 0.4 to 0.1),
$R^{(g)}$ its reward, $\bar R$ the mean reward of the item's $G$ draws, and $d$ the standard deviation of
the centred rewards in the minibatch.

Without $\bar R$ and $d$, $\hat g$ would be an unbiased estimate of the gradient of the expected reward
under noise, a smoothed version of $S$ that tends to $S$ as $\sigma\to0$. The two choices change this.
First, $\bar R$ contains the draw's own reward, so on average the estimate is shrunk by
$(G-1)/G=3/4$. Second, the reward differences shrink roughly in proportion to $\sigma$, so dividing by
$d$ makes the estimate grow roughly as $1/\sigma$ while the noise is annealed. Because CE keeps a fixed
weight, the noisy reward term gradually dominates it. Appendix~\ref{app:proof} states this formally
(Proposition~\ref{prop:rlcd}) and proves it, and Appendix~\ref{app:probe} measures it.

To remove these biases, use the mean of the \emph{other} $G-1$ rewards as the baseline and do not divide
by $d$. We call this \textbf{PG-LOO} (leave-one-out). We compare six objectives with identical data and
schedule: \textbf{CE}; \textbf{Proper}, which maximises $S$ without noise; \textbf{PG}, the released RLCD
recipe; \textbf{PG w/o CE}, its reward term alone; \textbf{PG-LOO}; and \textbf{Reparam}, which differentiates the smoothed
reward directly through the noise (the reparameterisation trick) and adds CE. PG-LOO and Reparam
estimate the same gradient on the same scale and differ only in variance.

\subsection{Calibration and long states}
\label{sec:calib}
\paragraph{Temperatures.} After training, one temperature per type and option-count bucket
($K=2$, $3$--$5$, $6$--$10$, ${\ge}11$) is fitted by minimising negative log-likelihood on a calibration
split carved from the training data before training, with a per-type fallback for buckets with fewer
than ten items. Every metric is reported before and after this step.

\paragraph{Long states.} Clinical notes exceed the 512-token context of base encoders. Besides
truncation and sliding windows, we evaluate \emph{retrieve-then-decide}: the model's own embedding
$e_\theta$ selects the $m=3$ note chunks most similar to the instruction $x$, and the decision is made on
those chunks in document order. This requires the converted model to remain a retriever.

%% file: sections/data.tex
\section{Data}
\label{sec:data}

\paragraph{\bench{}.} We convert existing biomedical datasets into typed decisions over a state
(Table~\ref{tab:data}; conversions in Appendix~\ref{app:tasks}). In the \textbf{grounded} track the
state determines the answer: evidence QA (PubMedQA; \citealp{jin2019pubmedqa}), claim verification
(SciFact, HealthVer and PUBHEALTH; \citealp{wadden2020fact,sarrouti2021evidence,kotonya2020explainable}),
drug--drug interactions (DDI; \citealp{herrero2013ddi}), cancer hallmarks with ten \noul{} questions per
abstract (HoC; \citealp{baker2016automatic}), adverse drug events (ADE; \citealp{gurulingappa2012development}),
patient drug reviews with three \score{} questions each (Druglib; \citealp{grasser2018aspect}), sentence
similarity (BIOSSES; \citealp{sogancioglu2017biosses}) and specialty routing (MTSamples). In the
\textbf{knowledge} track (MedQA, MedMCQA and MMLU-medical;
\citealp{jin2021disease,pal2022medmcqa,hendrycks2021measuring}) the state is an exam vignette whose
answer depends on memorised knowledge; it is reported but does not drive headline numbers. PUBHEALTH,
BIOSSES and MMLU are never trained on, and MTSamples is a held-out task family. We also evaluate on the
general-domain typed-decisions set used by Laya.
\input{tables/data_stats}

\paragraph{Conversion.} Each source example becomes one record: a \emph{state} (the text the decision is
about), one or more typed questions, and a gold distribution over each question's options
(Figure~\ref{fig:overview}; a full record is in Figure~\ref{fig:record}). The dataset's question or label set
becomes the instruction, and each label gets a short description, written once per task, so that options
carry content rather than bare label words. Multi-label annotations become one \noul{} per label (HoC),
multi-aspect ratings become several questions about one state (Druglib), a graded score becomes a \score{}
question whose target splits its mass between the two nearest levels (BIOSSES), and exam items become
\choice{} questions with the answer texts as descriptions. Answer-revealing fields are removed (PubMedQA
conclusions, PUBHEALTH explanations). Official splits are kept where they exist, and before any training a
calibration slice (about 10\% of training, at most 400 states) is carved off, grouped by claim or topic
where the source groups items. Appendix~\ref{app:tasks} details every dataset.

\paragraph{\medline{}.} Existing corpora pose one question per text, whereas System One requests
typically ask several heterogeneous questions about one state. Multi-question training data is usually
produced by an LLM teacher, whose errors then become targets. We instead derive labels from MEDLINE
indexing, which NLM curates: \emph{study design} (six classes from publication types, with a fixed
precedence for overlapping tags); \emph{disease area} (the single MeSH disease-tree category among major
topics, against 3--7 randomly drawn distractors, so that no encoder is favoured by model-mined negatives);
\emph{human or animal subjects}; whether \emph{adverse effects} or \emph{drug therapy} are reported; and,
for human studies, two sampled \emph{age or sex groups}. We keep 2005--2021 citations with manual or
human-curated indexing and exclude every PMID in any \bench{} evaluation set. Tags are weak labels:
absence need not mean false, and indexers may use the full article. The training set has 40k abstracts
(242,986 decisions); the test set has 5k abstracts from disjoint files.

\paragraph{Clinical track.} Under PhysioNet credentialed access~\citep{pollard2026physionet} we convert
MedNLI~\citep{romanov2018lessons,shivade2019mednli}; 1,000 human-reviewed trial-eligibility questions over
MIMIC-III discharge notes~\citep{woo2024synthetic,woo2025trialq,johnson2016mimic}, posed as
\emph{yes}/\emph{no}/\emph{not stated} and as an answerability \noul{}; and a name-mention \noul{} over
windows of MIMIC-IV notes with inserted surrogate names~\citep{deidnames2023,johnson2023mimiciv},
including a split of names that are also common words. The MIMIC-III notes average 12.9k characters and
form our long-state test. Credentialed data were processed only on institutional hardware and never sent
to a hosted API; the released model is trained on public data only.

%% file: tables/data_stats.tex
\begin{table}[t]
\caption{\bench{} and \medline{}, in decisions (a state may carry several questions).
$\dagger$~test only (never trained on); $\ddagger$~held-out task family; $\S$~training only.
Historical calibration slices are carved from the training data before training and
are not counted. Clinical training counts apply only to the +clinical variant. Counts describe the original splits; the integrity audit reports exclusions separately.}\label{tab:data}
\centering\footnotesize
\setlength{\tabcolsep}{2.5pt}
\begin{tabular}{@{}lllrr@{}}
\toprule
Task & Family & Type & \#Train & \#Test \\
\midrule
\multicolumn{5}{@{}l}{\textit{Grounded}} \\
PubMedQA-L & Evidence QA & \texttt{choice} & 400 & 500 \\
PubMedQA-A$^\S$ & Evidence QA & \texttt{noul} & 20,000 & -- \\
SciFact & Claim verif. & \texttt{choice} & 780 & 321 \\
HealthVer & Claim verif. & \texttt{choice} & 10,073 & 1,823 \\
PUBHEALTH$^\dagger$ & Claim verif. & \texttt{choice} & -- & 1,231 \\
DDI-2013 & Relation & \texttt{choice} & 18,379 & 7,039 \\
HoC & Doc.\ class. & \texttt{noul} & 11,660 & 3,710 \\
ADE & Adv.\ events & \texttt{noul} & 16,318 & 2,089 \\
Druglib & Review & \texttt{score} & 8,391 & 3,108 \\
BIOSSES$^\dagger$ & Similarity & \texttt{score} & -- & 100 \\
MTSamples$^\ddagger$ & Routing & \texttt{choice} & -- & 370 \\
\medline{} & Indexing & \texttt{mixed} & 242,986 & 30,522 \\
\multicolumn{5}{@{}l}{\textit{Knowledge}} \\
MedQA-USMLE & Exam & \texttt{choice} & 9,778 & 1,273 \\
MedMCQA & Exam & \texttt{choice} & 20,000 & 2,816 \\
MMLU-medical$^\dagger$ & Exam & \texttt{choice} & -- & 1,089 \\
\multicolumn{5}{@{}l}{\textit{General domain}} \\
typed-decisions$^\dagger$ & General & \texttt{mixed} & -- & 2,000 \\
\multicolumn{5}{@{}l}{\textit{Clinical (PhysioNet credentialed)}} \\
MedNLI & NLI & \texttt{choice} & 10,832 & 1,422 \\
MIMIC-III trials$^\dagger$ & Long notes & \texttt{mixed} & -- & 1,532 \\
MIMIC-IV names & Privacy & \texttt{noul} & 1,828 & 9,808 \\
\midrule
Total & & & 371,425 & 70,753 \\
\bottomrule
\end{tabular}
\end{table}

%% file: sections/setup.tex
\section{Experimental Setup}\label{sec:setup}
\paragraph{Encoders.} We evaluate eleven base-size checkpoints: PubMedBERT and its retrieval children
S-PubMedBERT-MS-MARCO and MedCPT-Query; MPNet and all-MPNet-base-v2; ModernBERT-base and its Nomic and GTE
embedders; BioClinical-ModernBERT and its embedder; and BGE-base-v1.5 as an unpaired
retriever~\citep{gu2021domain,deka2022improved,jin2023medcpt,song2020mpnet,allmpnet,warner2024modernbert,nussbaum2024nomic,zhang2024mgte,li2023towards,sounack2025bioclinicalmodernbertstateoftheartlongcontext,bcmbembed,xiao2024c}.
This gives six parent-retriever comparisons with shared parents and different retrieval corpora; they
are not six independent replications of one intervention. Model-specific query/document prefixes are used.

\paragraph{Training.} Training tasks are sampled with probability $\propto n_t^{1/2}$. The budget is 8,000
steps at batch size 32 (at most eight epochs), with AdamW, bf16 autocast over fp32 weights and gradient
clipping at norm 1. Options are shuffled except for ordinal rubrics. All encoders are trained with C and
\pfr{} under the released RLCD recipe (PG), three seeds each, plus 2\% and 10\% learning curves for the ten
paired checkpoints, also with three seeds. On S-PubMedBERT-MS-MARCO we run the
full \emph{head $\times$ objective} grid: C and \pfr{} with CE, Proper, PG, PG-LOO and Reparam (and PG w/o CE
for \pfr{}), three seeds per cell, identical data, schedule and seeds. Appendix~\ref{app:compute} lists
hyperparameters.

\paragraph{Baselines.} Laya-large (421M) is evaluated zero-shot and fine-tuned on our mixture with the
same recipe. Qwen3.8-27B~\citep{qwen38} and
Gemma-4-31B~\citep{gemmateam2026gemma4} score option letters in one
forward pass with thinking disabled (Appendix~\ref{app:prompt}). All systems are temperature-scaled on the
same calibration splits. Exam questions are likely present in LLM pre-training data, so LLM scores on the
knowledge track are descriptive only.

\paragraph{Metrics and statistics.} The headline metric is chance-normalised accuracy
$\mathrm{acc}_{\text{cn}}=(\mathrm{acc}-1/K)/(1-1/K)$, averaged over (question, $K$) groups within a task and
then over tasks. \emph{Seen} denotes the eight grounded training tasks; \emph{held-out} PUBHEALTH, BIOSSES and
MTSamples. We also report top-label ECE and NLL before and after temperature scaling, ordinal error,
option-order flips and retrieval retention (nDCG@10 on BEIR SciFact and
NFCorpus;~\citealp{thakur2021beir,boteva2016full}). Confidence intervals come from a paired bootstrap
(10,000 draws) over document clusters within each task: all records sharing a normalised text are
resampled together. Intervals condition on the observed tasks and on seed-averaged correctness; seed sd is
reported separately. Holm correction is applied within each research question across both tracks
(Appendix~\ref{app:metrics}). Non-significance is not read as equivalence.

\paragraph{Data integrity.} An exact-input audit found a few training duplicates in calibration (nine DDI
and seven Druglib records) and test (two SciFact and 13 Druglib records) splits, and repeated inputs within
some test sets. Removing all affected test inputs lowers seen-task scores by 0.3--0.4 points for every model
and changes no ranking (Appendix~\ref{app:integrity}).

%% file: sections/results.tex
\section{Results}\label{sec:results}
\input{tables/main_results}
\begin{figure*}[t]\centering
\includegraphics[width=\textwidth]{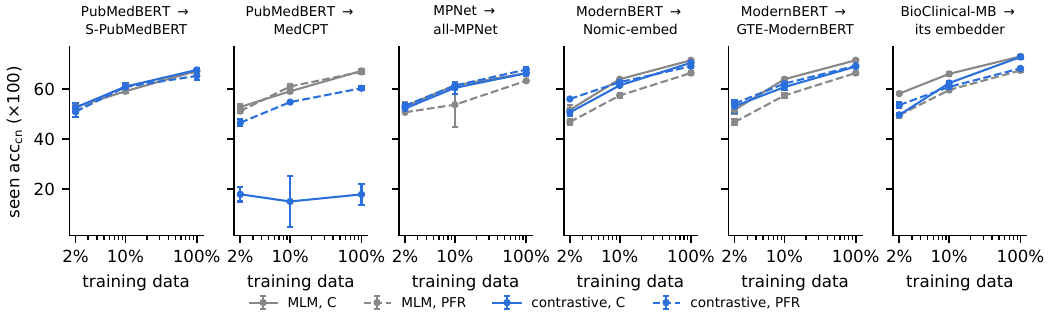}
\caption{Seen-task accuracy versus training data for the six parent-retriever comparisons (grey: MLM parent,
blue: retrieval child; solid: C, dashed: \pfr{}). Error bars show the sd over three seeds.}
\label{fig:lc}\end{figure*}

\subsection{RQ1: Does retrieval training help?}\label{sec:rq1}
\textbf{Retrieval improves zero-shot content matching.} Without training, every contrastive checkpoint reaches
15--24 chance-normalised points on questions whose options differ in content (disease areas, study
designs, specialties, rubrics, exam answers), against 0--7 for their MLM parents and 20.3 for Laya-large
(Table~\ref{tab:zeroshot}). On options that differ only by a label word (\emph{yes/no}, \emph{true/false}),
cosine gaps are of order $10^{-3}$ and no encoder is reliably above chance.

\textbf{After fine-tuning, retrieval benefits depend on the head.} With full data and the cross head the contrastive
stage is roughly neutral: S-PubMedBERT and PubMedBERT reach 67.7 and 67.3, both MPNet checkpoints 66.4, and
the BioClinical-ModernBERT pair 72.9 and 73.1 (Table~\ref{tab:main}); the only Holm-significant
difference is negative (GTE-ModernBERT $-2.5$; Table~\ref{tab:contrasts_rq1}). With \pfr{}, which keeps
the retrieval prior, the retriever is better in three of five pairs (all-MPNet $+4.5$, Nomic $+2.7$, GTE
$+2.9$), worse for S-PubMedBERT ($-1.6$), and the ModernBERT embedders gain $+10.9$ on held-out tasks
(Holm $p\le.014$ for each). With less data the heads diverge further (Figure~\ref{fig:lc}). For \pfr{},
retrievers are ahead in four of five pairs at both 10\% and 2\% (significant in three and four; e.g.\
Nomic $+5.3$ and $+9.2$, GTE $+4.7$ and $+7.4$, and up to $+16.3$ on held-out tasks). For C, the
ModernBERT-family retrievers are 2.5--3.7 points behind their parents at 10\%, BioClinical-ModernBERT's
embedder is 8.5 points behind at 2\%, and the only significant gain is S-PubMedBERT at 10\% ($+1.9$). With full training data, MedCPT-Query reaches
$17.9\pm4.3$ seen-task points with the cross head and $60.4\pm0.7$ with \pfr{} (sd over three seeds). In the tested
mask-slot setup the cross head's option logits have zero within-question spread in the saved predictions on 57\%,
86\% and 61\% of the 49,112 seen-task test questions in the three seeds (largest spread 0.0015), against about
0.1\% for \pfr{}. The \pfr{} prior uses MedCPT-Query to
encode both the state and the option hypotheses, whereas native MedCPT uses separate query and article
encoders. We report this pair descriptively in Table~\ref{tab:main} and Figure~\ref{fig:lc}; the formal RQ1
contrasts cover the other five pairs. \pfr{} improves on the cross head here but still trails its PubMedBERT
\pfr{} parent (66.9) by 6.6 points (computed from unrounded means). The strongest base-size model is
BioClinical-ModernBERT with a cross head (73.1), close to fine-tuned Laya-large (75.0) at about a third
of its size. Over all 15 pair and data-size comparisons (Table~\ref{tab:contrasts_rq1}; five pairs, three data
fractions, seen-task accuracy, Holm-adjusted $p<.05$), retrieval
training is significantly better in ten and worse in one with \pfr{}, and better in one and worse in five
with C.
\input{tables/zeroshot}

\subsection{RQ2: Which conversion works best?}\label{sec:rq2}
\input{tables/objectives}
\textbf{The cross head achieves higher accuracy.} In the matched grid (Table~\ref{tab:objectives}),
C exceeds \pfr{} by 1.3--2.4 seen-task points for all five shared objectives (Holm $p\le.011$ for CE,
Proper and PG; $.08$ and $.14$ for PG-LOO and Reparam) and by 3.8--10.2 points on held-out tasks. The
bi-encoder B matches C on seen tasks ($-0.6$, n.s.) but loses 9.4 held-out points; its value is that state
embeddings can be precomputed for corpus screening. A frozen prior improves \pfr{} by 2.5 points.

\textbf{The advantage of C is not uniform across tasks.} For S-PubMedBERT trained with CE, \pfr{} is higher than C
on HealthVer ($+6.9$) and Druglib ($+4.1$), whereas C is higher on SciFact ($+20.7$) and on the held-out MTSamples
($+19.4$; Table~\ref{tab:pertask}, seed-averaged, no intervals).

\textbf{Prior fusion reduces option-order sensitivity.} Prior fusion makes decisions almost invariant to option order: averaged over
all evaluation splits, a random permutation changes 0.3\% of S-PubMedBERT \pfr{} answers versus 15.7\% for
C, whose flips concentrate on unseen inputs (held-out 17\%, exams 28\%, seen tasks 3\%;
Appendix~\ref{app:robust}). The frozen-prior variant flips 13.1\% of its answers, compared with 0.3\% for the shared-prior variant (Table~\ref{tab:robust}). \pfr{} also has lower ordinal error on Druglib for every objective
(Table~\ref{tab:objfull}).

\input{tables/efficiency}
\textbf{Fine-tuning reduces retrieval performance.} SciFact nDCG@10 of the pooled embedding falls from 67.6 (Z) to 43.3
(C), 35.6 (\pfr{}), 34.3 (B) and 24.6 for C trained with CE (Table~\ref{tab:eff}); only a frozen prior
keeps the retriever intact. A ten-question request takes 6.6\,ms for C, 15.2\,ms for \pfr{} and 22.7\,ms
for Laya-large (forward pass, batch 1, MI300X).

\subsection{Frozen-prior ablation}\label{sec:frozen-ablation}
We tested whether the alignment between cosine scores and options accounts for the benefit
of the retrieval prior. We fine-tuned Nomic ModernBERT with cross-entropy on a fixed 2\%
sample from each of the cleaned public training tasks (2,592 states and 6,958 decisions).
Each condition used the same initialisation within a seed, the same training sample and
3,479 steps at batch size 16. The calibration and test splits were disjoint from training
and from each other by normalised state text. All nine runs used an NVIDIA L4 in BF16.
We compared a frozen cosine prior at fixed scale 20, the same scores reassigned using a
fixed, label-independent permutation within each question, and the same trainable
residual with no cosine scores. The permutation preserves the score distribution of
each question and is defined in canonical option order. Tables report seed means and
sample standard deviations over seeds 0--2.

\begin{table}[htbp]
\centering
\caption{Frozen-prior ablation. Task-macro chance-normalised accuracy in points, mean
$\pm$ seed standard deviation on 48,955 seen-task and 1,672 held-out decisions.}
\label{tab:frozen-ablation}
\small
\setlength{\tabcolsep}{4pt}
\begin{tabular}{lcc}
\toprule
Condition & Seen & Held out \\
\midrule
Real frozen prior & $48.04\pm1.18$ & $21.23\pm0.83$ \\
Shuffled frozen prior & $48.71\pm1.12$ & $28.60\pm2.50$ \\
Residual only & $50.49\pm1.19$ & $26.15\pm2.53$ \\
\bottomrule
\end{tabular}
\end{table}

The real-minus-shuffled difference was $-0.67\pm1.82$ points on seen tasks and
$-7.38\pm1.82$ on held-out tasks (mean $\pm$ standard deviation of paired seed differences).
The real prior also trailed the residual alone in every seed, with paired differences of
$-2.45\pm2.36$ and $-4.92\pm1.92$ points, respectively
(Table~\ref{tab:frozen-ablation}). PUBHEALTH contributed the largest held-out gap between
real and shuffled scores. These comparisons concern a single encoder, a fixed scale,
one data sample and the cross-entropy objective.

\subsection{RQ3: Does RLCD improve accuracy or calibration?}\label{sec:rq3}
\textbf{The released RLCD recipe trails cross-entropy.} With either head, CE, Proper and Reparam are within one point
of each other, while the released PG recipe trails CE by 3.0 (\pfr{}) and 2.5 (C) points (Holm
$p=.005$), with a fourfold larger seed sd for \pfr{} (1.5 vs.\ 0.4). Adding the spherical and RPS terms does
not significantly improve accuracy (Proper $-$ CE $+0.2$ and $+0.5$, n.s.), and removing CE from PG has little effect on accuracy ($+0.2$).

\begin{figure*}[tp]\centering
\includegraphics[width=0.92\textwidth]{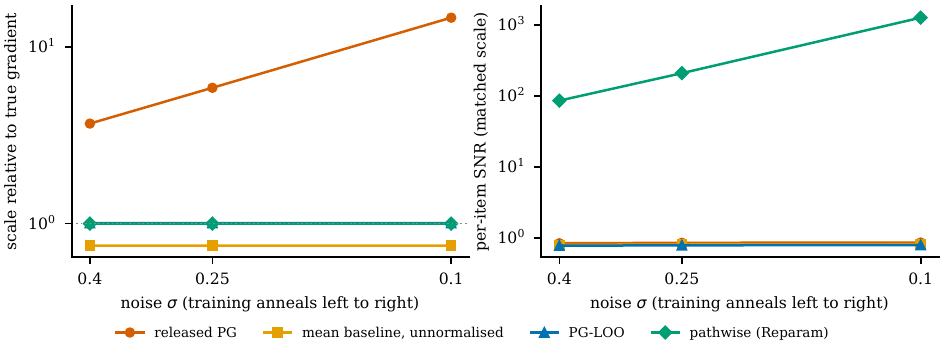}
\caption{Matched probe, averaged over the three checkpoints of Table~\ref{tab:probe}. Left: scale of each
estimator relative to the true smoothed-score gradient (dotted line: correct scale). Right: per-item
signal-to-noise ratio after matching scales. The released estimator's scale grows as $\sigma$ anneals; all
score-function estimators have SNR below one, the pathwise estimator two to three orders of magnitude more.}
\label{fig:probe}\end{figure*}
\textbf{An unbiased estimator reduces the accuracy deficit.} PG-LOO, which differs from PG only in its baseline and in
dropping the reward normalisation, gains 2.5 (\pfr{}) and 1.5 (C) points over PG (Holm $p=.005$) and is
within 0.8 points of Reparam (n.s.). A matched probe on fixed checkpoints and batches helps explain the gap
(Figure~\ref{fig:probe} and Appendix~\ref{app:probe}): the released estimator points in the right direction (cosine 0.98 with the
true smoothed gradient) but is 3.6--3.8, 5.7--6.1 and 14--15 times too large at $\sigma=0.4$, 0.25 and 0.1, so the
score term outweighs CE by 5--20$\times$ instead of 1.3$\times$. Every logged pre-clipping gradient norm
(one every 50 steps, all 33 grid runs) exceeds the clipping threshold of 1, and reward normalisation inflates the
score-function term relative to cross-entropy, which can change the direction of the gradient passed to AdamW;
under PG that direction is dominated by a score-function term with a per-item signal-to-noise ratio below one. The pathwise estimator is 100--1,500 times less noisy at matched scale,
yet PG-LOO trains nearly as well, so estimator variance is a secondary factor.

\textbf{No clear calibration benefit after scaling.} PG has the lowest raw ECE (7.0 vs.\ 9.3 for CE with
\pfr{}), but after temperature scaling all objectives lie between 3.6 and 4.4 and PG has the worst NLL
(Table~\ref{tab:objfull}, Figure~\ref{fig:reliability}). On the three held-out tasks the ordering is noisier
and differs from seen tasks (e.g.\ C with Reparam is 7 points below C with PG), so we do not generalise the
objective ranking beyond the training distribution.

\subsection{Calibration sensitivity to document overlap}\label{sec:clean-calibration}
We refitted temperature scaling for the three existing S-PubMedBERT cross-head
checkpoints trained with cross-entropy. We compared fitting on the original calibration
split with fitting after excluding states whose normalised text appeared in training or
the evaluated test tasks, and after removing repeated input records within each calibration
task. Both fits used the same regenerated calibration logits before filtering and were
scored on the same test records, after excluding training-document overlaps and repeated
inputs within each test task. The original-split refit reproduced historical task-macro
negative log likelihood (NLL) within 0.00006 for every seed and task group.

\begin{table}[htbp]
\centering
\caption{Calibration sensitivity on the retained test records. Task-macro means over three
seeds for negative log likelihood (NLL), Brier score and expected calibration error (ECE).}
\label{tab:clean-calibration}
\small
\setlength{\tabcolsep}{4pt}
\begin{tabular}{llrrr}
\toprule
Tasks & Temperature fit & NLL & Brier & ECE (\%) \\
\midrule
Seen & Original split & 0.5492 & 0.3030 & 5.81 \\
     & Filtered split & 0.5602 & 0.3057 & 6.11 \\
Held out & Original split & 1.3979 & 0.6802 & 17.49 \\
         & Filtered split & 1.3821 & 0.6721 & 15.84 \\
\bottomrule
\end{tabular}
\end{table}

The paired change in task-macro NLL was $+0.0110\pm0.0022$ on seen tasks and
$-0.0158\pm0.0089$ on held-out tasks (mean $\pm$ seed standard deviation).
On seen tasks the filtered refit still improved NLL over unscaled predictions
(0.6473 to 0.5602). On held-out tasks, NLL improved on MTSamples and worsened on
PUBHEALTH. Filtering retained 2,906 of 3,494 calibration records and 17,916 of 19,880
test records, including 100 of 1,823 HealthVer and 131 of 321 SciFact test records.
Table~\ref{tab:clean-calibration} therefore describes sensitivity on the retained records.

\subsection{LLMs and confidence-based routing}\label{sec:llm}
\begin{figure}[t]\centering
\includegraphics[width=\columnwidth]{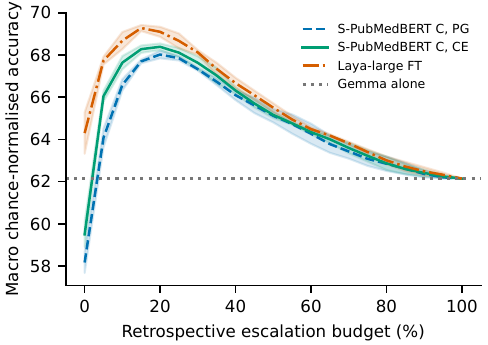}
\caption{Retrospective routing to Gemma-4-31B: the least-confident fraction of questions (by
temperature-scaled confidence) is escalated. Lines: mean over three encoder seeds; bands: sd.}
\label{fig:cascade}\end{figure}

Fine-tuned encoders are competitive on seen tasks: S-PubMedBERT C trained with CE (70.2) and Laya-large
(75.0) exceed zero-shot Gemma-4-31B (65.7) and Qwen3.8-27B (57.4), with better calibration (ECE after
scaling 3.8--4.3 vs.\ 8.5--9.1). The LLMs achieve higher accuracy on held-out tasks (Gemma 52.6 vs.\ 31.0), the
clinical track (81.6--83.4 vs.\ at most 38.4 zero-shot, 55.2 with clinical training) and exams. By task
(Table~\ref{tab:pertask}), both LLMs exceed every encoder shown there on PubMedQA, SciFact, PUBHEALTH and MTSamples, while
every encoder shown there exceeds both LLMs on DDI, HoC, ADE and MEDLINE-S1. This is consistent with the encoders
profiting from label schemes that the training mixture covers, an explanation we did not test. The two
are complementary (Figure~\ref{fig:cascade}): routing the 20\% least confident questions to Gemma raises
macro accuracy over seen and held-out tasks from \AloneSCE{} to \CascadeSCE{} for S-PubMedBERT C (CE)
and from \AloneL{} to \CascadeL{} for Laya-large, against \AloneGemma{} for Gemma alone. This is a
retrospective fixed budget on the test set, not a validated deployment threshold.

%% file: tables/main_results.tex
\begin{table*}[t]
\caption{Main results ($\times100$; mean$\pm$sd over 3 seeds; 10\% also uses 3 seeds). Encoders are grouped by
matched pair (MLM parent, then contrastive children). acc$_{\text{cn}}$: chance-normalised accuracy,
averaged over question groups within a task and then over tasks. Seen: the 8 grounded training tasks.
Held-out: PUBHEALTH, BIOSSES, MTSamples. Clinical: MedNLI, MIMIC-III trial questions and MIMIC-IV name
gate, zero-shot except ``+clinical train''. ECE$_{\text{TS}}$: top-label ECE after temperature scaling
on seen tasks. All rows are trained with the released RLCD (PG) recipe except ``C trained with CE''. In the lower block the C columns hold the row's own model. Knowledge-track results,
raw ECE and all architectures per encoder are in Appendix~\ref{app:full}.}\label{tab:main}
\centering\small
\setlength{\tabcolsep}{4.5pt}
\fittable{\begin{tabular}{@{}llccccccc@{}}
\toprule
 & & \multicolumn{3}{c}{Seen tasks, acc$_{\text{cn}}$} & \multicolumn{2}{c}{Held-out, acc$_{\text{cn}}$} & Clinical & ECE$_{\text{TS}}$ \\
\cmidrule(lr){3-5}\cmidrule(lr){6-7}
Encoder & Type & C & \pfr{} & C, 10\% & C & \pfr{} & C & C \\
\midrule
PubMedBERT & MLM & 67.3\tiny{$\pm$0.7} & 66.9\tiny{$\pm$0.8} & 59.1\tiny{$\pm$0.6} & 29.5\tiny{$\pm$3.3} & 20.7\tiny{$\pm$0.6} & 20.7\tiny{$\pm$0.8} & 4.8\tiny{$\pm$0.9} \\
S-PubMedBERT-MS-MARCO & contr. & 67.7\tiny{$\pm$0.9} & 65.3\tiny{$\pm$1.5} & 61.0\tiny{$\pm$1.4} & 32.8\tiny{$\pm$1.5} & 23.6\tiny{$\pm$1.5} & 20.5\tiny{$\pm$1.8} & 3.9\tiny{$\pm$0.7} \\
MedCPT-Query & contr. & 17.9\tiny{$\pm$4.3} & 60.4\tiny{$\pm$0.7} & 15.0\tiny{$\pm$10.2} & 3.9\tiny{$\pm$0.5} & 23.6\tiny{$\pm$3.2} & 12.5\tiny{$\pm$3.4} & 14.0\tiny{$\pm$2.3} \\
\midrule
MPNet & MLM & 66.4\tiny{$\pm$0.9} & 63.3\tiny{$\pm$0.4} & 61.5\tiny{$\pm$0.5} & 30.2\tiny{$\pm$2.7} & 20.8\tiny{$\pm$2.5} & 22.7\tiny{$\pm$5.2} & 4.8\tiny{$\pm$0.4} \\
all-MPNet-base-v2 & contr. & 66.4\tiny{$\pm$0.9} & 67.8\tiny{$\pm$0.8} & 60.5\tiny{$\pm$2.5} & 27.5\tiny{$\pm$0.8} & 24.7\tiny{$\pm$2.3} & 13.7\tiny{$\pm$5.0} & 5.2\tiny{$\pm$0.3} \\
\midrule
ModernBERT-base & MLM & 71.6\tiny{$\pm$0.7} & 66.5\tiny{$\pm$0.9} & 64.0\tiny{$\pm$0.6} & 32.5\tiny{$\pm$1.4} & 21.4\tiny{$\pm$1.8} & 26.1\tiny{$\pm$2.5} & 4.3\tiny{$\pm$0.2} \\
modernbert-embed-base & contr. & 70.6\tiny{$\pm$1.0} & 69.2\tiny{$\pm$0.6} & 61.5\tiny{$\pm$0.4} & 35.0\tiny{$\pm$1.1} & 32.3\tiny{$\pm$4.1} & 13.4\tiny{$\pm$2.9} & 4.9\tiny{$\pm$0.2} \\
gte-modernbert-base & contr. & 69.1\tiny{$\pm$0.8} & 69.3\tiny{$\pm$0.2} & 60.8\tiny{$\pm$1.2} & 29.8\tiny{$\pm$1.7} & 32.3\tiny{$\pm$0.2} & 24.3\tiny{$\pm$1.8} & 4.5\tiny{$\pm$0.1} \\
\midrule
BioClinical-ModernBERT & MLM & 73.1\tiny{$\pm$0.5} & 67.5\tiny{$\pm$0.5} & 66.1\tiny{$\pm$0.8} & 31.8\tiny{$\pm$2.0} & 21.0\tiny{$\pm$3.6} & 32.2\tiny{$\pm$4.1} & 4.1\tiny{$\pm$0.2} \\
BioClinical-MB-embeddings & contr. & 72.9\tiny{$\pm$0.7} & 68.3\tiny{$\pm$0.0} & 62.4\tiny{$\pm$0.9} & 32.1\tiny{$\pm$4.8} & 22.5\tiny{$\pm$1.7} & 26.5\tiny{$\pm$1.2} & 4.6\tiny{$\pm$0.5} \\
\midrule
BGE-base-v1.5 (unpaired) & contr. & 60.6\tiny{$\pm$0.5} & 66.8\tiny{$\pm$0.3} & -- & 16.1\tiny{$\pm$3.2} & 30.7\tiny{$\pm$3.0} & 7.8\tiny{$\pm$15.5} & 5.7\tiny{$\pm$0.7} \\
\midrule
\multicolumn{2}{@{}l}{PubMedBERT, B} & 67.0\tiny{$\pm$0.5} &  & -- & 17.8\tiny{$\pm$6.5} &  & 23.6\tiny{$\pm$1.4} & 4.1\tiny{$\pm$0.1} \\
\multicolumn{2}{@{}l}{S-PubMedBERT, B} & 67.1\tiny{$\pm$0.5} &  & -- & 23.4\tiny{$\pm$1.4} &  & 25.9\tiny{$\pm$2.1} & 3.9\tiny{$\pm$0.2} \\
\multicolumn{2}{@{}l}{S-PubMedBERT, C trained with CE} & 70.2\tiny{$\pm$0.4} &  & -- & 31.0\tiny{$\pm$1.7} &  & 20.0\tiny{$\pm$4.1} & 3.8\tiny{$\pm$0.2} \\
\multicolumn{2}{@{}l}{S-PubMedBERT, \pfr{} frozen prior} & 67.8\tiny{$\pm$1.0} &  & -- & 26.8\tiny{$\pm$1.9} &  & 19.5\tiny{$\pm$5.3} & 4.3\tiny{$\pm$0.2} \\
\multicolumn{2}{@{}l}{S-PubMedBERT, \pfr{} +clinical train} & 66.3\tiny{$\pm$0.2} &  & -- & 20.4\tiny{$\pm$0.8} &  & 55.2\tiny{$\pm$0.6} & 3.9\tiny{$\pm$0.4} \\
\multicolumn{2}{@{}l}{Laya-large (421M), C fine-tuned} & 75.0\tiny{$\pm$0.4} &  & -- & 35.8\tiny{$\pm$2.7} &  & 38.4\tiny{$\pm$4.5} & 4.3\tiny{$\pm$0.3} \\
\midrule
\multicolumn{2}{@{}l}{Qwen3.8-27B, zero-shot} & 57.4 &  & -- & 46.2 &  & 81.6 & 8.5 \\
\multicolumn{2}{@{}l}{Gemma-4-31B, zero-shot} & 65.7 &  & -- & 52.6 &  & 83.4 & 9.1 \\
\bottomrule
\end{tabular}}
\end{table*}

%% file: tables/zeroshot.tex
\begin{table}[t]
\caption{Untrained (step-0) chance-normalised accuracy ($\times100$, after temperature scaling on the
calibration split) on questions whose options differ in \emph{content} (MEDLINE design and disease
area, MTSamples, exam answers, Druglib and BIOSSES rubrics) or only by a \emph{label} word
(\noul{}, PubMedQA, verification, DDI, MedNLI). C at step 0: 3 seeds.}\label{tab:zeroshot}
\centering\small
\setlength{\tabcolsep}{3.5pt}
\fittable{\begin{tabular}{@{}lcccc@{}}
\toprule
 & \multicolumn{2}{c}{Z = \pfr{} at step 0} & \multicolumn{2}{c}{C at step 0} \\
\cmidrule(lr){2-3}\cmidrule(lr){4-5}
Encoder & content & label & content & label \\
\midrule
PubMedBERT & 1.0 & 13.0 & 1.1 & 24.0 \\
S-PubMedBERT-MS-MARCO & 15.2 & $-$20.4 & 2.3 & $-$2.8 \\
MedCPT-Query & 19.8 & $-$20.2 & 3.5 & 17.6 \\
MPNet & 1.0 & 3.6 & $-$1.2 & 12.5 \\
all-MPNet-base-v2 & 21.2 & $-$1.2 & $-$1.2 & 12.1 \\
ModernBERT-base & 0.0 & 17.1 & 0.6 & $-$7.1 \\
modernbert-embed-base & 18.9 & $-$5.0 & 2.4 & $-$19.8 \\
gte-modernbert-base & 23.3 & $-$13.8 & 1.0 & 0.0 \\
BioClinical-ModernBERT & 6.5 & $-$15.1 & 2.6 & 26.5 \\
BioClinical-MB-embeddings & 18.6 & 5.9 & 5.0 & $-$23.1 \\
BGE-base-v1.5 (unpaired) & 23.9 & 11.8 & 1.1 & 14.5 \\
\midrule
Laya-large (Laya's own head) & 20.3 & 36.2 & -- & -- \\
\bottomrule
\end{tabular}}
\end{table}

%% file: tables/objectives.tex
\begin{table}[t]
\caption{Head $\times$ objective on S-PubMedBERT-MS-MARCO ($\times100$, mean over 3 seeds; sd and further
metrics in Table~\ref{tab:objfull}). PG is the released RLCD recipe, used for all other trained models.
PG-LOO replaces its same-sample mean baseline and reward normalisation with a leave-one-out baseline.
PG w/o CE was run for \pfr{} only.}\label{tab:objectives}
\centering\small
\setlength{\tabcolsep}{3pt}
\fittable{\begin{tabular}{@{}lcccccc@{}}
\toprule
 & \multicolumn{2}{c}{Seen acc$_{\text{cn}}$} & \multicolumn{2}{c}{Held-out acc$_{\text{cn}}$} & \multicolumn{2}{c}{ECE$_{\text{TS}}$} \\
\cmidrule(lr){2-3}\cmidrule(lr){4-5}\cmidrule(lr){6-7}
Objective & \pfr{} & C & \pfr{} & C & \pfr{} & C \\
\midrule
CE & 68.3 & 70.2 & 20.8 & 31.0 & 4.0 & 3.8 \\
Proper & 68.5 & 70.7 & 22.6 & 28.8 & 4.4 & 3.9 \\
PG (released RLCD) & 65.3 & 67.7 & 23.6 & 32.8 & 3.8 & 3.9 \\
PG w/o CE & 65.5 & -- & 20.3 & -- & 4.0 & -- \\
PG-LOO (unbiased) & 67.8 & 69.2 & 23.8 & 30.0 & 3.9 & 3.7 \\
Reparam & 68.5 & 69.8 & 21.9 & 25.8 & 3.7 & 3.6 \\
\bottomrule
\end{tabular}}
\end{table}

%% file: tables/efficiency.tex
\ifdefined\pairedappendixtables\else\begin{table}[!htbp]\fi
\caption{Retrieval retention of the pooled embedding after conversion (seed 0), and p50 latency of one
request with 10 questions at batch size 1 on an MI300X (base-encoder latencies measured on the
PubMedBERT architecture, which S-PubMedBERT shares). Times cover the model forward pass only, excluding tokenisation and host/device transfers.}\label{tab:eff}
\centering\small
\setlength{\tabcolsep}{4pt}
\fittable{\begin{tabular}{@{}lccc@{}}
\toprule
 & \multicolumn{2}{c}{nDCG@10 ($\times100$)} & Latency \\
\cmidrule(lr){2-3}
Model (S-PubMedBERT) & SciFact & NFCorpus & ms \\
\midrule
Z (no training) & 67.6 & 31.3 & 10.2 \\
B & 34.3 & 11.0 & 9.7 \\
C & 43.3 & 13.0 & 6.6 \\
C, trained with CE & 24.6 & 10.3 & 6.6 \\
\pfr{}, shared prior & 35.6 & 11.2 & 15.2 \\
\pfr{}, frozen prior & 67.6 & 31.3 & 15.2 \\
Laya-large C (421M) & -- & -- & 22.7 \\
\bottomrule
\end{tabular}}
\ifdefined\pairedappendixtables\else\end{table}\fi

%% file: sections/discussion.tex
\section{Discussion}\label{sec:discussion}
\paragraph{Retrieval benefits depend on the head.} Retrieval training
helps zero-shot and helps a prior-fused head, more so with scarce labels, but is close to neutral for a
cross head with full data and can cost accuracy with few labels. Retrieval initialisation is therefore
most beneficial when the conversion retains the retrieval prior (\pfr{}), particularly with scarce
labels; it offers no consistent accuracy gain for a cross head.

\paragraph{What the frozen-prior ablation establishes.} The parent--retriever comparisons
evaluate retrieval-trained initialisation with a prior-fused head. They do not isolate
the cosine scores from the representation used by the residual. In the Nomic ModernBERT
ablation (Section~\ref{sec:frozen-ablation}), the residual alone exceeded the real frozen
prior in all three seeds on both task groups. Correct score assignment therefore did
not explain a performance gain in this fixed-scale, 2\%-data cross-entropy setting.
The comparison leaves open the effects of adapting the scale or jointly training
the retriever under the original objectives.

\paragraph{Choosing a conversion.} For per-request decisions, we recommend a cross head with cross-entropy
(70.2 on S-PubMedBERT). BioClinical-ModernBERT with a cross head reaches 73.1 under PG. \pfr{} is
preferable when low option-order sensitivity is important or when the cross head's option logits are
nearly tied, as for MedCPT-Query, and B when a corpus must be screened against many questions. Every trained conversion
degrades the embedding as a retriever unless the prior is frozen.

\paragraph{Implications for RLCD implementations.} The probe suggests a cheap check for any score-function
recipe: compare the scale of the reward gradient with that of the supervised term. In our runs every logged
pre-clipping gradient norm exceeded the threshold and reward normalisation raised the score-function term relative to
cross-entropy, which can change the direction of the gradient passed to AdamW; this is the mechanism we propose
for the released recipe's accuracy deficit. A leave-one-out
baseline without minibatch normalisation fixes the scale using the draws already made.

\section{Conclusion}\label{sec:conclusion}
Biomedical sentence encoders trained for retrieval can be converted into System One models:
with a cross head they reach 66--73 seen-task points on the original splits
(MedCPT-Query excepted), close to fine-tuned Laya-large (75.0) at about a third of its
size. Whether retrieval training itself helps depends on the conversion: it helps \pfr{} in 10 of 15
comparisons, most clearly with little labelled data, but does not consistently help C. Plain cross-entropy
with temperature scaling matches or outperforms the released RLCD recipe, whose deficit our probe suggests is largely due to reward
normalisation; implementations of RLCD should use an independent baseline such as leave-one-out. Fast
encoders and LLMs show complementary strengths in the retrospective routing experiment.
For per-request accuracy, a cross head trained with cross-entropy and temperature scaling
is a practical starting point. A frozen prior preserves retrieval, but its contribution
to decision accuracy should be checked against the residual alone. The calibration
sensitivity analysis shows that the benefit of refitting depends on the test task.
Open questions include native MedCPT query/article controls, independent validation of
routing thresholds, and source- or patient-level audits before any clinical claim.

\section*{Limitations}
\paragraph{Scope.} Results cover base-size encoders, one fixed training budget and hyperparameters
inherited from Laya; objectives were not tuned separately, and the head $\times$ objective grid was run on
one encoder. The six parent-retriever comparisons share parents and differ in retrieval corpora, so they
are not independent replications. Intervals condition on
the observed tasks and seed-averaged correctness; no mixed-effects or equivalence analysis is claimed.

\paragraph{Data.} \medline{} labels are weak: indexing can omit tags and uses the full article. A few exact
training inputs recur in calibration and test splits. We report an exact-input test-score
sensitivity analysis and a stricter document-filtered refit on three fixed checkpoints.
The filtered evaluation substantially reduces HealthVer and SciFact, and refitting
does not remove overlap from the original training history. Text-level identity cannot establish
patient-level independence or exclude near-duplicates. Exam contamination of the LLMs is unmeasured.

\paragraph{Calibration and shift.} Low ECE on seen tasks does not imply calibration on new tasks: on
held-out tasks several models are poorly calibrated, and option-order sensitivity of cross heads grows
under shift. Thresholds should be validated on in-domain samples before use.

\paragraph{Clinical use.} The clinical tests are small, come from one hospital system and use surrogate
names; long-note strategies remain close to chance for base encoders. Nothing here is a validated
clinical device, and the released model is not trained on clinical notes.

\section*{Broader Impact Statement}
\paragraph{Intended benefit.} Calibrated one-pass decision models can answer routine, schema-constrained
questions about biomedical text at low cost (median forward-pass latency of
6.6\,ms for a ten-question request with the cross head, batch size 1; Table~\ref{tab:eff}), and can escalate uncertain cases to a larger model or a person.
\paragraph{Risks.} Confident errors could exclude relevant evidence or influence a clinical decision. Calibration
measured on the training tasks does not carry over to new tasks, option-order sensitivity of cross heads grows under shift,
and the \medline{} labels are weak (Limitations). The models also inherit the biases and gaps of their source data.
\paragraph{Intended scope and oversight.} Nothing here is a validated clinical device, and the released model is not
trained on clinical notes. Any deployment should validate confidence thresholds on in-domain samples and keep human
review for escalated and low-confidence decisions in health-related settings; confidence alone is not a sufficient
safeguard. This study analysed existing datasets and did not collect new data from human participants.

\section*{Ethics Statement}
Credentialed PhysioNet data (MedNLI, MIMIC-III and MIMIC-IV derivatives) were used under their data use
agreements, processed only on institutional infrastructure, and never sent to a hosted model or API; only
aggregate metrics leave that infrastructure, and examples in this paper are paraphrased. Public datasets
are used under their published licences or terms of use; for them we release conversion code and
per-record hashes rather than the data. For \medline{} we release PMIDs, questions and labels under
CC BY 4.0 without the abstracts, which a script retrieves from PubMed. The released model is trained only on public data and inherits the non-commercial licence of S-PubMedBERT-MS-MARCO.
Calibrated confidences can encourage automation; we recommend human review of escalated and
low-confidence decisions in any health-related deployment.

%% file: sections/acknowledgements.tex
\section*{Acknowledgements}
We thank the creators and maintainers of the publicly released datasets, pretrained encoders,
language-model checkpoints and open-source software used in this study, including Laya,
Sentence-Transformers, PyTorch and Hugging Face Transformers. We also acknowledge the National Library
of Medicine for PubMed and MeSH, and PhysioNet and the contributors to the clinical resources accessed
under their respective data use agreements.

Computational experiments used Kelvin2 GPU resources provided by Queen's University Belfast through
the Northern Ireland High Performance Computing (NI-HPC) service. The public-data
calibration refits and frozen-prior ablation used Google Colab.

%% file: sections/declarations.tex
\section*{Statements and Declarations}
\paragraph{Funding.} The author received no funding for this work.
\paragraph{Competing interests.} The author declares no competing interests.
\paragraph{Ethics and consent.} This study used existing datasets and did not recruit
participants or obtain new consent. Credentialed data were handled under the applicable
data use agreements as described in the Ethics Statement.
\paragraph{Data and code availability.} Code and dataset builders are available at
\url{https://github.com/pritamdeka/sbert2s1}. The \medline{} labels and PubMed identifiers
are available at \url{https://huggingface.co/datasets/pritamdeka/MEDLINE-S1}, and the
released public-data model at \url{https://huggingface.co/pritamdeka/S1-PubMedBERT}.
Credentialed PhysioNet records are available separately to approved users under the
original providers' terms.
\paragraph{Author contributions.} Pritam Deka designed the study, implemented the
methods, conducted the experiments and analyses, and wrote and revised the manuscript.
\paragraph{Generative AI use.} Claude (Anthropic), Gemini (Google) and Codex (OpenAI)
assisted in preparing the manuscript and code. The author is responsible for the content.

%% file: sections/appendix.tex
\section{Extended Related Work}
\label{app:related}
\paragraph{System One models.} Besides Laya, Decider~\citep{decider2026} reads option-letter logits from
Qwen3.5 decoders fine-tuned with cross-entropy and fitted temperatures. The Decision
Index~\citep{decisionindex2026} scores typed-decision engines on tasks in five areas; its only clinical task
is trial-report inference. Concurrent evaluations use Jev and Laya as decision layers for
penetration-testing agents~\citep{pentest2026}, for text annotation in computational social
science~\citep{css2026} and for coding crash narratives~\citep{crash2026}; they study decision flips under
natural distracting context~\citep{jevout2026} and teach a Laya encoder which missing information to ask
for~\citep{lavoir2026}.

\paragraph{Biomedical benchmarks and cascades.} BLURB~\citep{gu2021domain} evaluates task-specific heads,
not runtime schemas or calibration. Biomedical QA
benchmarks~\citep{jin2019pubmedqa,jin2021disease,pal2022medmcqa} are dominated by generative LLMs. Routing
easy inputs to a cheap model and hard ones to an expensive one is the idea behind LLM
cascades~\citep{chen2023frugalgpt}; a calibrated System One model is a natural first stage
(\S\ref{sec:llm}).

\section{Task Conversions}
\label{app:tasks}
\begin{table*}[!t]\centering\scriptsize
\setlength{\tabcolsep}{3pt}
\begin{tabular}{@{}p{1.6cm}p{2.6cm}p{4.3cm}p{3.1cm}p{3.3cm}@{}}
\toprule
Task & State & Question(s) and type & Options and target & Splits \\
\midrule
PubMedQA-L & abstract contexts without the conclusion & the original question (\choice{}) & yes / no / maybe with short descriptions; one-hot & 1,000 labelled items: 500 test, 50 dev, 50 calibration, rest train (seeded) \\
PubMedQA-A & abstract contexts & the original question (\noul{}) & yes $\to$ true; one-hot & 20k sampled for training, excluding every evaluation PMID \\
SciFact & title and abstract of the evidence (or first cited) document & ``Does this abstract support or refute the claim: \emph{c}'' (\choice{}) & supports / refutes / not\_enough\_info & official train; validation as test; calibration grouped by claim \\
HealthVer & evidence passage & ``Does this evidence support or refute the claim: \emph{c}'' (\choice{}) & as SciFact & official train / dev / test; calibration grouped by topic \\
PUBHEALTH & article text (fact-checkers' explanation removed) & ``Based on this article, what is the veracity of the health claim: \emph{c}'' (\choice{}) & false / mixture / true / unproven & official test only \\
DDI-2013 & sentence with the two drugs marked & ``What kind of interaction between @DRUG1\$ and @DRUG2\$ does the sentence describe?'' (\choice{}) & mechanism / effect / advice / int / none & official dev as test (test is gated); test sentences seen in training removed \\
HoC & abstract (sentences joined) & ``Does this abstract discuss the cancer hallmark `\emph{h}'?'', one \noul{} per hallmark & sentence labels aggregated per abstract & official train / dev / test \\
ADE & sentence & ``Does this sentence describe an adverse effect caused by a drug?'' (\noul{}) & one-hot & deduplicated; seeded 80/10/10 \\
Druglib & drug, condition, and the benefit, side-effect and comment reviews & effectiveness, side-effect severity and overall rating (three \score{}) & 5 levels each; the 1--10 rating binned in pairs & official train / test \\
BIOSSES & first sentence & ``How similar in meaning is this sentence to: \emph{s$_2$}'' (\score{}) & 5 levels; mass split between the two levels adjacent to the annotator mean & all 100 pairs as test \\
MTSamples & transcription & ``Which medical specialty does this transcription belong to?'' (\choice{}) & every specialty in the test set, without descriptions & test only \\
MedQA, MedMCQA, MMLU-med. & question stem & ``Which option correctly answers the question in the text?'' (\choice{}) & answer texts as descriptions & official splits; MedMCQA single-answer, 20k sampled for training \\
\medline{} & title and abstract & design and disease area (\choice{}); humans, animals, adverse effects, drug therapy, two age or sex groups (\noul{}) & from publication types and MeSH; 3--7 random disease distractors & disjoint baseline files; evaluation PMIDs excluded \\
\bottomrule
\end{tabular}
\caption{Conversion of each public source dataset into typed decisions. Every training set loses a calibration
slice of $\min(400,\max(10,\lfloor n/10\rfloor))$ states, carved before any run and used only for temperature
fitting. Clinical conversions are listed in the text below.}
\label{tab:conversions}
\end{table*}
Table~\ref{tab:conversions} lists the public conversions. The clinical tasks are converted in the same
way: MedNLI asks ``How does this hypothesis relate to the clinical note? Hypothesis: \ldots with
options \{entailment, neutral, contradiction\}; the trial questions are posed verbatim with
\{yes, no, not\_stated\}, plus the \noul{} ``Can this question be answered from the information in the
note: \ldots; and the name gate asks ``Does this passage mention a person's name (a patient, relative or
clinician)? over windows of MIMIC-IV notes.

\par\smallskip\noindent\begin{minipage}{\columnwidth}
\scriptsize
\begin{verbatim}
{"task": "medline_s1", "split": "test",
 "state": "<title and abstract of a PubMed citation>",
 "questions": {
  "design": {"type": "choice",
   "instructions": "What type of study or article is this?",
   "criteria": {
    "randomized_trial": "a randomized controlled trial",
    "case_report": "a case report of one or a few patients",
    "...": "(four more designs)"}},
  "humans": {"type": "noul", "instructions": "Did this
   research study human beings (patients or participants)?"}},
 "gold": {
  "design": {"probabilities": {"randomized_trial": 1.0}},
  "humans": {"probabilities": {"true": 1.0}}}}
\end{verbatim}
\vspace{-6pt}
\captionof{figure}{A \medline{} record in the shared request format (abridged). Gold labels are distributions over
option keys. A \score{} question lists its levels from low to high, and its keys are level indices.}
\label{fig:record}
\end{minipage}\par\smallskip

\section{RLCD: Formal Statement and Proof}\label{app:proof}
\paragraph{Definitions.} Let $t\in\Delta^{K-1}$ be the target and $p\in\Delta^{K-1}$ a reported
distribution. Laya's reward is
\begin{equation}
\begin{split}
  S(p,t)={}&\sum_k t_k\log p_k+w_{\mathrm{sph}}\frac{\langle t,p\rangle}{\lVert p\rVert_2}\\
  &-w_{\mathrm{rps}}\,\ind[\tau{=}\score]\,\mathrm{RPS}(p,t),
\end{split}
\label{eq:score}
\end{equation}
with $\mathrm{RPS}(p,t)=\frac{1}{K-1}\sum_{k}\big(\sum_{j\le k}(p_j-t_j)\big)^2$, $w_{\mathrm{sph}}=0.75$,
$w_{\mathrm{rps}}=1$, and the log term floored at $\log\delta$, $\delta>0$. Each step draws $G=4$
perturbations $\varepsilon^{(g)}\sim\mathcal{N}(0,\sigma^2\Pi_K)$, where $\Pi_K=I_K-\tfrac1K\mathbf{1}\mathbf{1}^\top$
restricts noise to directions that change the softmax. Each perturbed report $\softmax(z+\varepsilon^{(g)})$
receives reward $R^{(g)}=S(\cdot,t)$, and RLCD minimises
\begin{equation}
\begin{split}
  \mathcal{L}_{\mathrm{PG}}={}&-\frac1G\sum_{g}A^{(g)}\log\pi_z\big(z{+}\varepsilon^{(g)}\big)\\
  &+\lambda\,\mathrm{CE}(z,t),
\end{split}
\label{eq:pgloss}
\end{equation}
where $\pi_z=\mathcal{N}(z,\sigma^2\Pi_K)$, $\lambda=1$ ($\lambda=0$ for PG w/o CE), and the advantage is
$A^{(g)}=(R^{(g)}-\bar R)/d$. Since $\nabla_z\log\pi_z(z+\varepsilon)=\varepsilon/\sigma^2$ on the
centred subspace, the reward part of $-\nabla_z\mathcal{L}_{\mathrm{PG}}$ is $\hat g$ in Eq.~\eqref{eq:pg}.

\begin{proposition}\label{prop:rlcd}
Let $J_\sigma(z)=\E_{\varepsilon\sim\mathcal{N}(0,\sigma^2\Pi_K)}\big[S(\softmax(z+\varepsilon),t)\big]$. Then
\begin{enumerate}[label=(\roman*),leftmargin=*,itemsep=1pt,topsep=2pt]
  \item $\nabla_z J_\sigma=\sigma^{-2}\,\E[(S(\softmax(z{+}\varepsilon),t)-b)\,\varepsilon]$ for any $b$ independent
  of $\varepsilon$ (score-function form);
  \item $\nabla_z J_\sigma=\E\big[\nabla_z S(\softmax(z{+}\varepsilon),t)\big]$ (pathwise form);
  \item $J_\sigma(z)\to S(\softmax(z),t)$ as $\sigma\to0$, which equals $-\mathrm{CE}(z,t)$ for
  $w_{\mathrm{sph}}=w_{\mathrm{rps}}=0$ wherever the floor is inactive;
  \item with the same-sample mean $\bar R$ as baseline and no normalisation, the score-function estimate
  has expectation $\tfrac{G-1}{G}\nabla_z J_\sigma$.
\end{enumerate}
\end{proposition}
\noindent RLCD therefore performs stochastic ascent on a noise-smoothed proper score. PG-LOO uses
$b_{-g}=\frac{1}{G-1}\sum_{h\ne g}R^{(h)}$ and no normalisation, which by (i) is unbiased and on the same
scale as the pathwise estimator (ii) used by Reparam.

\paragraph{Proof.}
Let $V=\{v:\mathbf1^\top v=0\}$. Softmax is invariant to shifts along $\mathbf1$, so we may centre $z$ onto
$V$; $\varepsilon\sim\mathcal{N}(0,\sigma^2\Pi_K)$ is an isotropic Gaussian on $V$ with density
$\varphi_\sigma$. Write $f(y)=S(\softmax(y),t)$. With the log term floored at $\log\delta$, the spherical term
in $[0,1]$ and RPS in $[0,1]$, $f$ is bounded and Lipschitz, hence differentiable almost everywhere.

\emph{(i)} $J_\sigma(z)=\int_V f(y)\varphi_\sigma(y-z)\,dy$; differentiating under the integral (justified by
boundedness of $f$ and integrability of $\nabla\varphi_\sigma$) gives, for directions in $V$,
\begin{align*}
\nabla J_\sigma(z)&=\int_V f(y)\,\frac{y-z}{\sigma^2}\,\varphi_\sigma(y-z)\,dy\\
&=\sigma^{-2}\,\E[f(z+\varepsilon)\,\varepsilon].
\end{align*}
Any $b$ independent of $\varepsilon$ satisfies $\E[b\,\varepsilon]=\E[b]\,\E[\varepsilon]=0$.

\emph{(ii)} $f$ is Lipschitz, so $\nabla f$ exists almost everywhere and is bounded; dominated convergence
gives $\nabla\E[f(z+\varepsilon)]=\E[\nabla f(z+\varepsilon)]$.

\emph{(iii)} $f$ is continuous and bounded and $\varepsilon\to0$ in distribution, so $J_\sigma(z)\to f(z)$. With
$w_{\mathrm{sph}}=w_{\mathrm{rps}}=0$ and an inactive floor, $f(z)=\sum_k t_k\log\softmax(z)_k=-\mathrm{CE}(z,t)$.

\emph{(iv)} For $G$ independent draws with $R_g=f(z+\varepsilon_g)$ and $\bar R=G^{-1}\sum_h R_h$,
\begin{align*}
\E\big[(R_g-\bar R)\,\varepsilon_g\big]&=\Big(1-\tfrac1G\Big)\E[R_g\varepsilon_g]\\
&\quad-\tfrac1G\sum_{h\ne g}\E[R_h]\,\E[\varepsilon_g],
\end{align*}
and the last term vanishes, so the averaged estimate has expectation $\frac{G-1}{G}\nabla J_\sigma$.
With the leave-one-out baseline $b_{-g}=(G-1)^{-1}\sum_{h\ne g}R_h$, which is independent of
$\varepsilon_g$, the expectation is exactly $\nabla J_\sigma$ by (i). \hfill$\square$

\paragraph{Normalisation.} The released recipe further divides all advantages by the standard deviation
$d$ of the centred rewards in the minibatch. Because $d$ depends on the same draws, the factor cannot be
taken outside the expectation; more importantly, the centred rewards and hence $d$ shrink roughly in proportion to $\sigma$, while the
estimate carries a factor $\varepsilon/\sigma^2$, so the effective scale of the score term grows roughly as
$1/\sigma$ as $\sigma$ is annealed. Since the CE term enters on a fixed scale, the normalisation changes the relative
weight of the two terms during training. Figure~\ref{fig:probe} shows how the scale varies with noise.

\section{Gradient-Estimator Probe and Optimisation Traces}\label{app:probe}
\input{tables/probe}
\begin{figure*}[!t]\centering
\includegraphics[width=0.92\textwidth]{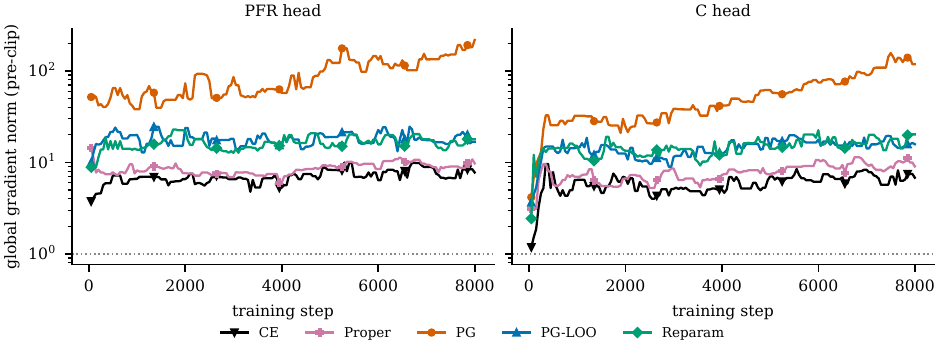}
\caption{Median global gradient norm before clipping during training (S-PubMedBERT-MS-MARCO, three seeds,
rolling median). The clipping threshold is 1 (dotted); every logged norm (one every 50 steps) in every run of the grid exceeds it.
The released PG recipe's norm is an order of magnitude above CE and grows as
$\sigma$ anneals.}
\label{fig:gradnorm}\end{figure*}
The probe (Table~\ref{tab:probe}, Figure~\ref{fig:probe}) holds the model, batch and noise level fixed and
compares estimators of the gradient of the smoothed score with respect to the logits. Three results are
consistent across a \pfr{} checkpoint trained with PG, one trained with PG-LOO, and a C checkpoint trained
with CE. First, the same-sample mean baseline yields a slope of exactly $0.75=(G-1)/G$, confirming
Proposition~\ref{prop:rlcd}(iv); PG-LOO and the pathwise estimator have slope 1. Second, the released
normalisation multiplies the estimate by 3.6--3.8 at $\sigma=0.4$ and by 14--15 at $\sigma=0.1$, so relative to
cross-entropy the score term weighs 4.7--5.0 and then 19--20 times as much, against 1.3 for the unbiased
estimators. Third, at matched scale all score-function estimators have per-item SNR around 0.8, whereas the
pathwise estimator's SNR is 85--1,300. Figure~\ref{fig:gradnorm} shows the consequence during training:
every logged gradient norm exceeds the clipping threshold for every objective, and under PG the direction
passed to AdamW is dominated by the inflated, noisy score term. Because PG-LOO trains almost as well as Reparam despite its far higher variance, we
attribute the released recipe's deficit primarily to this reweighting.

\section{Complete Head $\times$ Objective Results}\label{app:objfull}
\input{tables/objectives_full}
\begin{figure*}[tp]\centering
\includegraphics[width=\textwidth]{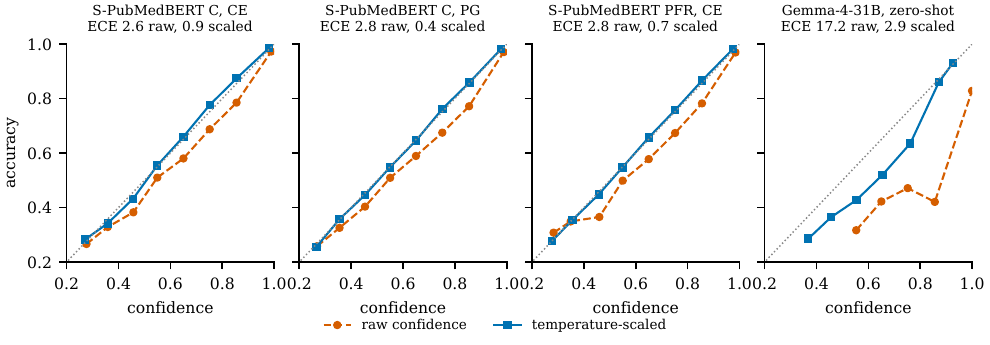}
\caption{Top-label reliability diagrams on the seen tasks (items pooled over tasks and seeds; 10 bins,
bins with fewer than 30 items omitted), before (dashed) and after (solid) temperature scaling. Pooled
ECE differs from the task-averaged ECE in the tables. Fine-tuned encoders are over-confident before scaling
and close to the diagonal after it; the zero-shot LLM remains miscalibrated.}
\label{fig:reliability}\end{figure*}
Table~\ref{tab:objfull} reports every metric of the matched grid. Three patterns complement
\S\ref{sec:rq3}. (1)~The head effect is consistent: C has higher seen and held-out accuracy and lower NLL
for every objective, \pfr{} lower ordinal error and option-order sensitivity. (2)~Raw ECE is lowest for the
score-function variants (PG, PG w/o CE, PG-LOO), not for Reparam, which optimises the same smoothed
objective; temperature scaling removes the difference (Figure~\ref{fig:reliability}). (3)~The pre-clip gradient norm orders the objectives
consistently (CE $<$ Proper $<$ PG-LOO $\approx$ Reparam $\ll$ PG).

\section{Metrics and Statistics}\label{app:metrics}
Accuracy and top-label ECE use the argmax target; ordinal MAE uses the expected predicted level against
it. BIOSSES has soft targets, so hard-label summaries discard part of its annotation. ECE uses 15
equal-width bins; saved metrics also include equal-mass and class-wise ECE, Brier score, AURC, AUROC, AUPRC
and ordinal metrics.

Paired bootstrap intervals use 10,000 draws of document clusters with replacement within each task: text is
whitespace-normalised and hashed, and all records and questions that share a text are resampled together.
Task weights stay fixed, and correctness is averaged over seeds before resampling, so intervals omit seed
uncertainty and are pointwise. Two-sided $p$ values use $2\min(\Pr[\Delta^*\le0],\Pr[\Delta^*\ge0])$ with a
$+1$ correction (smallest attainable value $2/10{,}001$). Holm correction is applied within four families (RQ1:
60 tests; RQ2: 14; RQ3: 26; baselines: 8), across both tracks. Tables~\ref{tab:contrasts_rq1}
and~\ref{tab:contrasts} list every contrast.
\input{tables/contrasts_rq1}
\input{tables/contrasts}

\section{Full Results by Encoder and Task}\label{app:full}
\input{tables/main_full}
\input{tables/per_task}
Table~\ref{tab:full} lists every encoder with each trained architecture, including raw ECE and the knowledge
track, on which no base-size encoder exceeds 15 chance-normalised points. Table~\ref{tab:pertask} breaks the
seen and held-out tracks down by task. The cross head's advantage over \pfr{} is concentrated on tasks that
require comparing two texts in detail (SciFact: 55--61 vs.\ 35--40) and on held-out specialty routing
(MTSamples: 69--71 vs.\ 51 for S-PubMedBERT); on single-text classification (DDI, HoC, ADE, MEDLINE) the two heads are within
a few points.

\section{Robustness and Long Notes}\label{app:robust}
\input{tables/robustness}
\paragraph{Option order.} Averaged over all evaluation splits, cross heads change 10--17\% of their answers
under a random permutation of the options, including Laya-large; the flips concentrate on held-out, exam and
clinical inputs (S-PubMedBERT C: 3\% on seen tasks, 17\% held-out, 28\% on exams). B and \pfr{} with a shared
prior are essentially invariant, because the shared prior scores each option independently of its position;
with a frozen prior, \pfr{} flips 13\%, which we did not investigate further.

\paragraph{Long notes.} The MIMIC-III trial questions are small (532 answer and 1,000 answerability items)
and no strategy consistently performs best. For the three-way answer, sliding windows help most models: Laya-large
rises from 10.9 to 20.8 with the most confident window and S-PubMedBERT B from 10.9 to 20.5. For
answerability, retrieve-then-decide is best for S-PubMedBERT C (20.2) and its frozen-prior \pfr{} (29.9), but
for most other models all strategies are close to chance. Long-note decisions remain challenging for
the evaluated base encoders.

\paragraph{Privacy gate.} Names that are also common words have similar AUROC to ordinary names (the two
columns differ by at most two points). Training on the gate's own windows (+clinical) raises AUROC from
about 75 to 99.8; both LLMs are perfect zero-shot.

\section{Data Integrity and Sensitivity}\label{app:integrity}
An input identity comprises the whitespace-normalised state and question schema, excluding labels. Nine
DDI and seven Druglib calibration records, and two SciFact and 13 Druglib test records, have inputs that occur
in training. Repeated inputs also occur within some test sets (HealthVer, DDI, Druglib, MTSamples,
MMLU-medical). Identical evidence paired with different claims (HealthVer, SciFact) is task-native and is
not counted as duplication. Table~\ref{tab:sensitivity} re-scores the original checkpoints after excluding
every affected test input using the original temperatures. Section~\ref{sec:clean-calibration}
reports a separate temperature-refit analysis using normalized document identities and a common
filtered test subset.
\input{tables/sensitivity}

\section{LLM Prompt}
\label{app:prompt}
The prompt is rendered with each model's chat template, with thinking disabled:
\begin{quote}\small\ttfamily
Read the text and answer the question.\\[2pt]
Text:\\ \{state\}\\[2pt]
Question: \{instructions\}\\[2pt]
Options:\\ A. \{option 1\}\\ B. \{option 2\} \ldots\\[2pt]
Respond with the letter of the correct option only.
\end{quote}
Option probabilities are the softmax over the letters' next-token log-probabilities; for each letter the
bare and the space-prefixed token are combined by log-sum-exp. States longer than 6,000 tokens are truncated.

\section{Hyperparameters and Compute}
\label{app:compute}
\begin{itemize}[leftmargin=*,itemsep=1pt,topsep=2pt]
  \item \textbf{Sequence limits}: 512 tokens per cross sequence, of which at most 192 are question and option
  tokens (48 per option); bi-encoder states 256 tokens, hypotheses 64.
  \item \textbf{Optimisation}: AdamW; learning rates $2.5{\times}10^{-5}$ (encoder), $10^{-4}$ (head),
  $5{\times}10^{-2}$ ($\alpha$); weight decay 0.01 (none on $\alpha$); 5\% warm-up then cosine decay; global
  gradient-norm clipping at 1; bf16 autocast over fp32 weights.
  \item \textbf{Objectives}: $G=4$; $\sigma$ annealed linearly from 0.4 to 0.1; $\delta=10^{-4}$;
  $w_{\mathrm{sph}}=0.75$, $w_{\mathrm{rps}}=1$, $\lambda=1$ (0 for PG w/o CE).
  \item \textbf{Temperatures}: fitted by L-BFGS on $\log T$, clipped to $[0.05,20]$; buckets with fewer than 10
  calibration items fall back to the per-type value.
  \item \textbf{Hardware and packing}: AMD MI300X (192\,GB) GPUs; a shared file queue packs up to six runs per
  GPU, each capped at a declared memory budget (22\,GB for base encoders, 44\,GB for Laya-large, 100--110\,GB
  for the LLMs). Runs checkpoint every 20 minutes and on pre-emption and resume exactly, since the data
  schedule is a function of seed and step.
  \item \textbf{Throughput} (single run, batch 32): C 265 items/s (11.7\,GB); \pfr{} 223 items/s (16.3\,GB);
  B 326 items/s (6.4\,GB); Laya-large C 99 items/s (36.3\,GB).
  \item \textbf{Compute}: the initial campaign (125 runs including both LLM baselines, plus 8 smoke tests)
  took 11.9 wall-clock hours on three GPUs (about 36 GPU-hours; 170 task-hours under packing). The
  head $\times$ objective controls and gradient probe (16 tasks) took 1.3 wall-clock hours on three GPUs
  (14.6 task-hours). The 50 additional learning-curve runs for the ModernBERT and BioClinical pairs took
  4.2 wall-clock hours on four GPUs (72 task-hours). The ModernBERT/BGE full-data extension and the GTE
  reruns used further allocations that we did not account separately.
\end{itemize}

\section{Released Software and Model}\label{app:release}
The code release contains the four conversions, the five objectives (including PG-LOO), temperature
fitting, the long-state strategies, the data builders, a script that restores the \medline{} abstracts
from PubMed, and the evaluation and bootstrap code used here.
The released model is the best seed (by seen-task accuracy) of S-PubMedBERT-MS-MARCO with a cross head
trained with CE on public data (seen 70.5; three-seed mean $70.2\pm0.4$). Inference is a single file of
plain PyTorch (\texttt{sbert2s1.py}) that ships with the model. The Hugging Face example below downloads
and loads this file directly; dependencies are PyTorch, Transformers, Safetensors and Hugging Face Hub:
\par\smallskip
\noindent\begin{minipage}{\columnwidth}\footnotesize\ttfamily\raggedright
import importlib.util as iu\\
from huggingface\_hub import hf\_hub\_download\\
repo = "\ifpreprint pritamdeka/S1-PubMedBERT\fi"\\
path = hf\_hub\_download(repo, "sbert2s1.py")\\
spec = iu.spec\_from\_file\_location("s1", path)\\
s1 = iu.module\_from\_spec(spec)\\
spec.loader.exec\_module(s1)\\
m = s1.load(repo)\\
abstract = "Participants were randomised."\\
q = \{"design": \{"type": "choice",\\
\hspace*{2em}"instructions": "Study design?",\\
\hspace*{2em}"criteria": \{"rct": "randomised trial",\\
\hspace*{6em}"cohort": "cohort study"\}\}\}\\
m.predict(abstract, q)
\end{minipage}\par\smallskip
\noindent\begin{minipage}{\columnwidth}
Answers follow the Jev/Laya format with the fitted temperatures applied, and truncated inputs are marked.
The checkpoint is also exported in Laya's layout. The model inherits the CC BY-NC 2.0 licence of its
base encoder; the code is released under the Apache 2.0 licence.
\ifpreprint Code: \url{https://github.com/pritamdeka/sbert2s1}; model:
\url{https://huggingface.co/pritamdeka/S1-PubMedBERT}; data:
\url{https://huggingface.co/datasets/pritamdeka/MEDLINE-S1}.
\fi
\end{minipage}\par

%% file: tables/probe.tex
\begin{table*}[!t]
\caption{Matched gradient-estimator probe. For each saved checkpoint, the logits of the same 40 training batches
(32 items each) are fixed, and 32 independent estimates of the smoothed-score gradient ($G{=}4$, no CE term)
are drawn per estimator and noise level. Slope: projection of the mean estimate onto a 2{,}048-sample pathwise
reference $\nabla J_\sigma$ (1 is the correct scale). Score : CE weight: slope $\times\lVert\nabla J_\sigma\rVert/\lVert\nabla\mathrm{CE}\rVert$,
the effective weight of the smoothed-score term relative to the CE term in a training step. SNR: per-item
$\lVert\nabla J_\sigma\rVert^2$ divided by the scale-matched variance of one estimate.}\label{tab:probe}
\centering\small
\setlength{\tabcolsep}{3pt}
\fittable{\begin{tabular}{@{}llccccccc@{}}
\toprule
 & & \multicolumn{3}{c}{Slope vs.\ $\nabla J_\sigma$} & \multicolumn{2}{c}{Score : CE weight} & Cosine & SNR \\
\cmidrule(lr){3-5}\cmidrule(lr){6-7}
Checkpoint (head, objective) & Estimator & $\sigma{=}0.4$ & $0.25$ & $0.1$ & $\sigma{=}0.4$ & $0.1$ & $\sigma{=}0.25$ & $\sigma{=}0.25$ \\
\midrule
\pfr{}, PG & released PG & 3.59 & 5.72 & 14.27 & 4.74 & 18.87 & 0.982 & 0.861 \\
 & mean baseline, unnormalised & 0.75 & 0.75 & 0.75 & 0.99 & 1.00 & 0.982 & 0.816 \\
 & PG-LOO & 1.01 & 1.00 & 1.00 & 1.33 & 1.33 & 0.983 & 0.791 \\
 & pathwise (Reparam) & 1.00 & 1.00 & 1.00 & 1.32 & 1.32 & 1.000 & 206 \\
\midrule
\pfr{}, PG-LOO & released PG & 3.66 & 5.84 & 14.57 & 4.83 & 19.23 & 0.982 & 0.854 \\
 & mean baseline, unnormalised & 0.75 & 0.75 & 0.75 & 0.99 & 0.99 & 0.982 & 0.805 \\
 & PG-LOO & 1.01 & 1.00 & 1.00 & 1.32 & 1.32 & 0.983 & 0.792 \\
 & pathwise (Reparam) & 1.00 & 1.00 & 1.00 & 1.32 & 1.32 & 1.000 & 209 \\
\midrule
C, CE & released PG & 3.80 & 6.05 & 15.09 & 5.02 & 19.97 & 0.982 & 0.847 \\
 & mean baseline, unnormalised & 0.75 & 0.75 & 0.75 & 0.99 & 0.99 & 0.982 & 0.804 \\
 & PG-LOO & 1.01 & 1.01 & 1.01 & 1.33 & 1.33 & 0.984 & 0.781 \\
 & pathwise (Reparam) & 1.00 & 1.00 & 1.00 & 1.32 & 1.32 & 1.000 & 216 \\
\bottomrule
\end{tabular}}
\end{table*}

%% file: tables/objectives_full.tex
\begin{table*}[tp]
\caption{Complete head $\times$ objective grid on S-PubMedBERT-MS-MARCO (mean$\pm$sd over 3 seeds; accuracy,
ECE and flip rates $\times100$). Columns 3--6: chance-normalised accuracy by track. ECE, NLL: seen tasks, top-label,
before and after temperature scaling. MAE$_{\score}$: error of the expected Druglib level. Perm.\ flip: share of
seen-task answers that change under a random option permutation. $\lVert\nabla\rVert$: median global gradient norm
before clipping; the clipping threshold is 1 and every logged norm of every objective exceeds it.}\label{tab:objfull}
\centering\small
\setlength{\tabcolsep}{3pt}
\fittable{\begin{tabular}{@{}llcccccccccc@{}}
\toprule
Objective & Head & Seen & Held-out & Clinical & Knowledge & ECE & ECE$_{\text{TS}}$ & NLL$_{\text{TS}}$ & MAE$_{\score}$ & Perm.\ flip & $\lVert\nabla\rVert$ \\
\midrule
CE & \pfr{} & 68.3\tiny{$\pm$0.4} & 20.8\tiny{$\pm$1.6} & 21.9\tiny{$\pm$2.9} & 13.1\tiny{$\pm$0.8} & 9.3\tiny{$\pm$0.9} & 4.0\tiny{$\pm$0.4} & 0.510\tiny{$\pm$0.004} & 0.588\tiny{$\pm$0.012} & 0.4\tiny{$\pm$0.1} & 7.1 \\
 & C & 70.2\tiny{$\pm$0.4} & 31.0\tiny{$\pm$1.7} & 20.0\tiny{$\pm$4.1} & 13.6\tiny{$\pm$0.9} & 7.4\tiny{$\pm$0.3} & 3.8\tiny{$\pm$0.2} & 0.491\tiny{$\pm$0.001} & 0.639\tiny{$\pm$0.020} & 2.9\tiny{$\pm$0.4} & 6.0 \\
\midrule
Proper & \pfr{} & 68.5\tiny{$\pm$0.2} & 22.6\tiny{$\pm$1.9} & 23.4\tiny{$\pm$0.7} & 13.6\tiny{$\pm$0.4} & 9.1\tiny{$\pm$0.6} & 4.4\tiny{$\pm$0.4} & 0.510\tiny{$\pm$0.006} & 0.582\tiny{$\pm$0.004} & 0.2\tiny{$\pm$0.1} & 8.7 \\
 & C & 70.7\tiny{$\pm$0.3} & 28.8\tiny{$\pm$0.4} & 22.6\tiny{$\pm$1.1} & 12.8\tiny{$\pm$0.7} & 7.6\tiny{$\pm$0.5} & 3.9\tiny{$\pm$0.3} & 0.493\tiny{$\pm$0.013} & 0.627\tiny{$\pm$0.007} & 2.6\tiny{$\pm$0.2} & 7.7 \\
\midrule
PG (released RLCD) & \pfr{} & 65.3\tiny{$\pm$1.5} & 23.6\tiny{$\pm$1.5} & 18.3\tiny{$\pm$8.2} & 11.4\tiny{$\pm$1.2} & 7.0\tiny{$\pm$0.6} & 3.8\tiny{$\pm$0.2} & 0.548\tiny{$\pm$0.018} & 0.638\tiny{$\pm$0.016} & 0.4\tiny{$\pm$0.1} & 88.1 \\
 & C & 67.7\tiny{$\pm$0.9} & 32.8\tiny{$\pm$1.5} & 20.5\tiny{$\pm$1.8} & 11.2\tiny{$\pm$0.7} & 6.5\tiny{$\pm$0.0} & 3.9\tiny{$\pm$0.7} & 0.527\tiny{$\pm$0.010} & 0.746\tiny{$\pm$0.031} & 3.2\tiny{$\pm$0.5} & 50.2 \\
\midrule
PG w/o CE & \pfr{} & 65.5\tiny{$\pm$2.0} & 20.3\tiny{$\pm$3.3} & 16.9\tiny{$\pm$11.0} & 10.6\tiny{$\pm$0.9} & 6.6\tiny{$\pm$0.4} & 4.0\tiny{$\pm$0.4} & 0.543\tiny{$\pm$0.014} & 0.646\tiny{$\pm$0.019} & 0.2\tiny{$\pm$0.2} & 76.2 \\
 & C & \multicolumn{10}{c}{not run} \\
\midrule
PG-LOO (unbiased) & \pfr{} & 67.8\tiny{$\pm$0.6} & 23.8\tiny{$\pm$5.7} & 25.2\tiny{$\pm$3.5} & 12.4\tiny{$\pm$0.2} & 7.5\tiny{$\pm$0.5} & 3.9\tiny{$\pm$0.2} & 0.518\tiny{$\pm$0.009} & 0.605\tiny{$\pm$0.014} & 0.4\tiny{$\pm$0.1} & 17.6 \\
 & C & 69.2\tiny{$\pm$0.6} & 30.0\tiny{$\pm$2.1} & 20.5\tiny{$\pm$0.9} & 11.7\tiny{$\pm$0.3} & 6.6\tiny{$\pm$0.4} & 3.7\tiny{$\pm$0.5} & 0.509\tiny{$\pm$0.005} & 0.665\tiny{$\pm$0.001} & 3.3\tiny{$\pm$0.8} & 14.7 \\
\midrule
Reparam & \pfr{} & 68.5\tiny{$\pm$0.5} & 21.9\tiny{$\pm$2.0} & 20.5\tiny{$\pm$5.5} & 9.4\tiny{$\pm$6.2} & 9.0\tiny{$\pm$0.8} & 3.7\tiny{$\pm$0.4} & 0.508\tiny{$\pm$0.005} & 0.587\tiny{$\pm$0.009} & 0.4\tiny{$\pm$0.2} & 16.0 \\
 & C & 69.8\tiny{$\pm$1.6} & 25.8\tiny{$\pm$1.5} & 20.4\tiny{$\pm$1.2} & 11.4\tiny{$\pm$0.4} & 7.2\tiny{$\pm$0.6} & 3.6\tiny{$\pm$0.4} & 0.498\tiny{$\pm$0.021} & 0.668\tiny{$\pm$0.040} & 2.4\tiny{$\pm$0.1} & 14.4 \\
\bottomrule
\end{tabular}}
\end{table*}

%% file: tables/contrasts_rq1.tex
\begin{table*}[!t]
\caption{Initialisation contrasts (contrastive child minus MLM parent) for five of the six pairs (MedCPT-Query is omitted; see Section~\ref{sec:rq1}), both heads and every training-data fraction. Document-clustered paired bootstrap (10,000 draws; differences in chance-normalised points). Intervals are pointwise and condition on the observed tasks and on seed-averaged correctness. Holm adjustment is applied within each family (RQ1; RQ2; RQ3; baselines) across both tracks. The smallest attainable unadjusted $p$ is 0.0002.}\label{tab:contrasts_rq1}
\centering\small
\renewcommand{\arraystretch}{0.85}
\setlength{\tabcolsep}{4pt}
\fittable{\begin{tabular}{@{}lccccc@{}}
\toprule
Contrast ($a - b$) & Seeds & Seen $\Delta$ [95\% CI] & $p_{\text{Holm}}$ & Held-out $\Delta$ [95\% CI] & $p_{\text{Holm}}$ \\
\midrule
C full: S-PubMedBERT $-$ PubMedBERT & 3/3 & +0.4 [$-$0.4, +1.1] & 1 & +3.2 [+0.3, +6.1] & .956 \\
C full: all-MPNet $-$ MPNet & 3/3 & +0.0 [$-$0.8, +0.9] & 1 & $-$2.6 [$-$5.6, +0.3] & 1 \\
C full: Nomic-embed $-$ ModernBERT & 3/3 & $-$1.0 [$-$1.9, $-$0.1] & .708 & +2.5 [$-$1.2, +6.2] & 1 \\
C full: GTE-ModernBERT $-$ ModernBERT & 3/3 & $-$2.5 [$-$3.5, $-$1.5] & .012 & $-$2.7 [$-$5.9, +0.6] & 1 \\
C full: BioClinical-MB-emb $-$ BioClinical-MB & 3/3 & $-$0.2 [$-$1.0, +0.5] & 1 & +0.3 [$-$3.2, +3.7] & 1 \\
PFR full: S-PubMedBERT $-$ PubMedBERT & 3/3 & $-$1.6 [$-$2.5, $-$0.8] & .014 & +2.9 [$-$0.1, +6.1] & 1 \\
PFR full: all-MPNet $-$ MPNet & 3/3 & +4.5 [+3.5, +5.5] & .012 & +3.9 [+0.1, +7.7] & 1 \\
PFR full: Nomic-embed $-$ ModernBERT & 3/3 & +2.7 [+1.7, +3.7] & .012 & +10.9 [+6.7, +15.2] & .012 \\
PFR full: GTE-ModernBERT $-$ ModernBERT & 3/3 & +2.9 [+1.9, +3.9] & .012 & +10.9 [+6.4, +15.3] & .012 \\
PFR full: BioClinical-MB-emb $-$ BioClinical-MB & 3/3 & +0.8 [$-$0.1, +1.7] & 1 & +1.5 [$-$1.4, +4.5] & 1 \\
C 2\%: S-PubMedBERT $-$ PubMedBERT & 3/3 & $-$0.2 [$-$1.1, +0.7] & 1 & +0.5 [$-$1.9, +2.8] & 1 \\
C 2\%: all-MPNet $-$ MPNet & 3/3 & $-$0.8 [$-$1.8, +0.2] & 1 & +12.3 [+9.7, +14.9] & .012 \\
C 2\%: Nomic-embed $-$ ModernBERT & 3/3 & $-$1.0 [$-$2.0, 0.0] & 1 & +3.1 [+0.3, +5.8] & .918 \\
C 2\%: GTE-ModernBERT $-$ ModernBERT & 3/3 & +1.5 [+0.5, +2.5] & .112 & $-$4.3 [$-$7.5, $-$1.2] & .243 \\
C 2\%: BioClinical-MB-emb $-$ BioClinical-MB & 3/3 & $-$8.5 [$-$9.7, $-$7.4] & .012 & +2.3 [$-$0.9, +5.5] & 1 \\
PFR 2\%: S-PubMedBERT $-$ PubMedBERT & 3/3 & $-$0.2 [$-$1.2, +0.7] & 1 & +7.6 [+4.1, +11.1] & .012 \\
PFR 2\%: all-MPNet $-$ MPNet & 3/3 & +2.6 [+1.5, +3.7] & .012 & +1.1 [$-$2.4, +4.5] & 1 \\
PFR 2\%: Nomic-embed $-$ ModernBERT & 3/3 & +9.2 [+8.1, +10.4] & .012 & +15.6 [+12.2, +19.1] & .012 \\
PFR 2\%: GTE-ModernBERT $-$ ModernBERT & 3/3 & +7.4 [+6.2, +8.6] & .012 & +15.5 [+11.4, +19.7] & .012 \\
PFR 2\%: BioClinical-MB-emb $-$ BioClinical-MB & 3/3 & +4.0 [+2.8, +5.2] & .012 & +0.6 [$-$2.9, +4.2] & 1 \\
C 10\%: S-PubMedBERT $-$ PubMedBERT & 3/3 & +1.9 [+0.9, +2.8] & .012 & $-$2.6 [$-$5.0, $-$0.4] & .654 \\
C 10\%: all-MPNet $-$ MPNet & 3/3 & $-$1.0 [$-$2.0, +0.0] & 1 & $-$3.5 [$-$6.7, $-$0.3] & .956 \\
C 10\%: Nomic-embed $-$ ModernBERT & 3/3 & $-$2.5 [$-$3.5, $-$1.5] & .012 & +6.6 [+3.1, +10.0] & .012 \\
C 10\%: GTE-ModernBERT $-$ ModernBERT & 3/3 & $-$3.1 [$-$4.2, $-$2.1] & .012 & +3.0 [$-$0.2, +6.2] & 1 \\
C 10\%: BioClinical-MB-emb $-$ BioClinical-MB & 3/3 & $-$3.7 [$-$4.6, $-$2.7] & .012 & +1.5 [$-$1.8, +5.0] & 1 \\
PFR 10\%: S-PubMedBERT $-$ PubMedBERT & 3/3 & $-$0.2 [$-$1.0, +0.6] & 1 & +0.1 [$-$3.2, +3.4] & 1 \\
PFR 10\%: all-MPNet $-$ MPNet & 3/3 & +7.7 [+6.6, +8.8] & .012 & +6.3 [+3.3, +9.4] & .012 \\
PFR 10\%: Nomic-embed $-$ ModernBERT & 3/3 & +5.3 [+4.2, +6.4] & .012 & +16.3 [+12.5, +20.1] & .012 \\
PFR 10\%: GTE-ModernBERT $-$ ModernBERT & 3/3 & +4.7 [+3.6, +5.9] & .012 & +13.5 [+9.4, +17.8] & .012 \\
PFR 10\%: BioClinical-MB-emb $-$ BioClinical-MB & 3/3 & +1.2 [+0.2, +2.2] & .558 & $-$3.1 [$-$6.6, +0.5] & 1 \\
\bottomrule
\end{tabular}}
\end{table*}

%% file: tables/contrasts.tex
\begin{table*}[!t]
\caption{Conversion (RQ2), objective (RQ3) and baseline contrasts on S-PubMedBERT-MS-MARCO. Document-clustered paired bootstrap (10,000 draws; differences in chance-normalised points). Intervals are pointwise and condition on the observed tasks and on seed-averaged correctness. Holm adjustment is applied within each family (RQ1; RQ2; RQ3; baselines) across both tracks. The smallest attainable unadjusted $p$ is 0.0002.}\label{tab:contrasts}
\centering\small
\renewcommand{\arraystretch}{0.85}
\setlength{\tabcolsep}{4pt}
\fittable{\begin{tabular}{@{}lccccc@{}}
\toprule
Contrast ($a - b$) & Seeds & Seen $\Delta$ [95\% CI] & $p_{\text{Holm}}$ & Held-out $\Delta$ [95\% CI] & $p_{\text{Holm}}$ \\
\midrule
C $-$ PFR (CE) & 3/3 & +1.9 [+0.8, +3.1] & .011 & +10.2 [+5.8, +14.6] & .003 \\
C $-$ PFR (Proper) & 3/3 & +2.2 [+1.1, +3.4] & .004 & +6.2 [+2.0, +10.5] & .041 \\
C $-$ PFR (PG) & 3/3 & +2.4 [+1.2, +3.6] & .003 & +9.2 [+5.0, +13.4] & .003 \\
C $-$ PFR (LOO) & 3/3 & +1.5 [+0.3, +2.6] & .079 & +6.2 [+2.0, +10.3] & .041 \\
C $-$ PFR (Reparam) & 3/3 & +1.3 [+0.1, +2.4] & .136 & +3.8 [$-$0.6, +8.2] & .247 \\
B $-$ C (PG) & 3/3 & $-$0.6 [$-$1.8, +0.7] & .353 & $-$9.3 [$-$13.5, $-$5.0] & .003 \\
PFR frozen $-$ shared prior (PG) & 3/3 & +2.5 [+1.3, +3.6] & .003 & +3.2 [$-$0.5, +7.0] & .247 \\
PFR: CE $-$ PG & 3/3 & +3.0 [+2.2, +3.8] & .005 & $-$2.8 [$-$5.7, +0.2] & .735 \\
PFR: Reparam $-$ PG & 3/3 & +3.2 [+2.5, +4.0] & .005 & $-$1.7 [$-$4.7, +1.5] & 1 \\
PFR: LOO $-$ PG & 3/3 & +2.5 [+1.7, +3.2] & .005 & +0.2 [$-$2.9, +3.3] & 1 \\
PFR: LOO $-$ Reparam & 3/3 & $-$0.8 [$-$1.4, $-$0.1] & .322 & +1.9 [$-$1.0, +4.9] & 1 \\
PFR: LOO $-$ CE & 3/3 & $-$0.5 [$-$1.1, +0.1] & .929 & +3.0 [+0.5, +5.5] & .259 \\
PFR: Proper $-$ CE & 3/3 & +0.2 [$-$0.3, +0.7] & 1 & +1.8 [$-$0.7, +4.4] & 1 \\
PFR: PG w/o CE $-$ PG & 3/3 & +0.2 [$-$0.5, +0.9] & 1 & $-$3.3 [$-$5.8, $-$0.9] & .139 \\
C: CE $-$ PG & 3/3 & +2.5 [+1.8, +3.2] & .005 & $-$1.8 [$-$3.8, +0.4] & .929 \\
C: Reparam $-$ PG & 3/3 & +2.1 [+1.4, +2.8] & .005 & $-$7.0 [$-$10.3, $-$3.8] & .005 \\
C: LOO $-$ PG & 3/3 & +1.5 [+0.9, +2.2] & .005 & $-$2.8 [$-$5.1, $-$0.5] & .259 \\
C: LOO $-$ Reparam & 3/3 & $-$0.6 [$-$1.2, +0.0] & .725 & +4.2 [+1.9, +6.6] & .015 \\
C: LOO $-$ CE & 3/3 & $-$1.0 [$-$1.6, $-$0.3] & .043 & $-$1.0 [$-$3.1, +1.1] & 1 \\
C: Proper $-$ CE & 3/3 & +0.5 [0.0, +1.1] & .735 & $-$2.2 [$-$4.5, +0.0] & .725 \\
Laya-large FT $-$ S-PubMedBERT C (PG) & 3/3 & +7.3 [+6.1, +8.5] & .002 & +3.0 [$-$0.5, +6.5] & .094 \\
Laya-large FT $-$ S-PubMedBERT C (CE) & 3/3 & +4.8 [+3.7, +5.9] & .002 & +4.8 [+1.2, +8.4] & .021 \\
Gemma-4-31B $-$ Laya-large FT & 1/3 & $-$9.3 [$-$10.7, $-$7.8] & .002 & +16.8 [+10.4, +23.1] & .002 \\
Gemma-4-31B $-$ S-PubMedBERT C (CE) & 1/3 & $-$4.5 [$-$6.0, $-$2.9] & .002 & +21.6 [+16.2, +26.9] & .002 \\
\bottomrule
\end{tabular}}
\end{table*}

%% file: tables/main_full.tex
\begin{table*}[!t]
\caption{All encoders and architectures ($\times100$; mean$\pm$sd over 3 seeds).}\label{tab:full}
\centering\small
\renewcommand{\arraystretch}{0.88}
\setlength{\tabcolsep}{4pt}
\fittable{\begin{tabular}{@{}llcccccc@{}}
\toprule
Encoder & Arch & Seen acc$_{\text{cn}}$ & ECE & ECE$_{\text{TS}}$ & Held-out & Clinical & Knowledge \\
\midrule
PubMedBERT & C & 67.3\tiny{$\pm$0.7} & 6.7\tiny{$\pm$0.6} & 4.8\tiny{$\pm$0.9} & 29.5\tiny{$\pm$3.3} & 20.7\tiny{$\pm$0.8} & 13.4\tiny{$\pm$0.7} \\
 & PFR & 66.9\tiny{$\pm$0.8} & 8.3\tiny{$\pm$0.4} & 4.1\tiny{$\pm$0.3} & 20.7\tiny{$\pm$0.6} & 23.9\tiny{$\pm$0.6} & 12.5\tiny{$\pm$0.9} \\
 & B & 67.0\tiny{$\pm$0.5} & 8.3\tiny{$\pm$0.4} & 4.1\tiny{$\pm$0.1} & 17.8\tiny{$\pm$6.5} & 23.6\tiny{$\pm$1.4} & 11.5\tiny{$\pm$0.7} \\
S-PubMedBERT-MS-MARCO & C & 67.7\tiny{$\pm$0.9} & 6.5\tiny{$\pm$0.0} & 3.9\tiny{$\pm$0.7} & 32.8\tiny{$\pm$1.5} & 20.5\tiny{$\pm$1.8} & 11.2\tiny{$\pm$0.7} \\
 & PFR & 65.3\tiny{$\pm$1.5} & 7.0\tiny{$\pm$0.6} & 3.8\tiny{$\pm$0.2} & 23.6\tiny{$\pm$1.5} & 18.3\tiny{$\pm$8.2} & 11.4\tiny{$\pm$1.2} \\
 & B & 67.1\tiny{$\pm$0.5} & 7.8\tiny{$\pm$0.6} & 3.9\tiny{$\pm$0.2} & 23.4\tiny{$\pm$1.4} & 25.9\tiny{$\pm$2.1} & 11.8\tiny{$\pm$0.7} \\
MedCPT-Query & C & 17.9\tiny{$\pm$4.3} & 14.0\tiny{$\pm$2.3} & 14.0\tiny{$\pm$2.3} & 3.9\tiny{$\pm$0.5} & 12.5\tiny{$\pm$3.4} & 2.8\tiny{$\pm$0.6} \\
 & PFR & 60.4\tiny{$\pm$0.7} & 7.1\tiny{$\pm$0.2} & 4.8\tiny{$\pm$0.3} & 23.6\tiny{$\pm$3.2} & 12.6\tiny{$\pm$3.7} & 10.1\tiny{$\pm$0.7} \\
\midrule
MPNet & C & 66.4\tiny{$\pm$0.9} & 8.2\tiny{$\pm$0.4} & 4.8\tiny{$\pm$0.4} & 30.2\tiny{$\pm$2.7} & 22.7\tiny{$\pm$5.2} & 4.6\tiny{$\pm$1.0} \\
 & PFR & 63.3\tiny{$\pm$0.4} & 7.6\tiny{$\pm$0.2} & 4.3\tiny{$\pm$0.4} & 20.8\tiny{$\pm$2.5} & 16.9\tiny{$\pm$1.6} & 5.7\tiny{$\pm$0.6} \\
all-MPNet-base-v2 & C & 66.4\tiny{$\pm$0.9} & 8.0\tiny{$\pm$0.3} & 5.2\tiny{$\pm$0.3} & 27.5\tiny{$\pm$0.8} & 13.7\tiny{$\pm$5.0} & 3.3\tiny{$\pm$0.7} \\
 & PFR & 67.8\tiny{$\pm$0.8} & 8.1\tiny{$\pm$0.3} & 4.0\tiny{$\pm$0.2} & 24.7\tiny{$\pm$2.3} & 18.4\tiny{$\pm$2.9} & 7.2\tiny{$\pm$0.4} \\
\midrule
ModernBERT-base & C & 71.6\tiny{$\pm$0.7} & 8.0\tiny{$\pm$0.5} & 4.3\tiny{$\pm$0.2} & 32.5\tiny{$\pm$1.4} & 26.1\tiny{$\pm$2.5} & 7.1\tiny{$\pm$0.8} \\
 & PFR & 66.5\tiny{$\pm$0.9} & 8.5\tiny{$\pm$0.5} & 4.4\tiny{$\pm$0.1} & 21.4\tiny{$\pm$1.8} & 19.9\tiny{$\pm$5.1} & 6.5\tiny{$\pm$0.4} \\
modernbert-embed-base & C & 70.6\tiny{$\pm$1.0} & 8.9\tiny{$\pm$0.3} & 4.9\tiny{$\pm$0.2} & 35.0\tiny{$\pm$1.1} & 13.4\tiny{$\pm$2.9} & 7.0\tiny{$\pm$1.2} \\
 & PFR & 69.2\tiny{$\pm$0.6} & 8.3\tiny{$\pm$0.3} & 4.1\tiny{$\pm$0.2} & 32.3\tiny{$\pm$4.1} & 12.4\tiny{$\pm$2.3} & 8.2\tiny{$\pm$0.2} \\
gte-modernbert-base & C & 69.1\tiny{$\pm$0.8} & 8.7\tiny{$\pm$0.5} & 4.5\tiny{$\pm$0.1} & 29.8\tiny{$\pm$1.7} & 24.3\tiny{$\pm$1.8} & 8.2\tiny{$\pm$1.3} \\
 & PFR & 69.3\tiny{$\pm$0.2} & 8.7\tiny{$\pm$0.3} & 4.0\tiny{$\pm$0.5} & 32.3\tiny{$\pm$0.2} & 23.2\tiny{$\pm$2.1} & 10.4\tiny{$\pm$0.5} \\
\midrule
BioClinical-ModernBERT & C & 73.1\tiny{$\pm$0.5} & 8.0\tiny{$\pm$0.5} & 4.1\tiny{$\pm$0.2} & 31.8\tiny{$\pm$2.0} & 32.2\tiny{$\pm$4.1} & 10.0\tiny{$\pm$0.7} \\
 & PFR & 67.5\tiny{$\pm$0.5} & 8.2\tiny{$\pm$0.3} & 4.2\tiny{$\pm$0.3} & 21.0\tiny{$\pm$3.6} & 17.3\tiny{$\pm$7.1} & 8.0\tiny{$\pm$1.1} \\
BioClinical-MB-embeddings & C & 72.9\tiny{$\pm$0.7} & 8.5\tiny{$\pm$0.5} & 4.6\tiny{$\pm$0.5} & 32.1\tiny{$\pm$4.8} & 26.5\tiny{$\pm$1.2} & 8.4\tiny{$\pm$2.3} \\
 & PFR & 68.3\tiny{$\pm$0.0} & 8.3\tiny{$\pm$0.1} & 4.3\tiny{$\pm$0.4} & 22.5\tiny{$\pm$1.7} & 6.2\tiny{$\pm$4.8} & 7.8\tiny{$\pm$0.1} \\
\midrule
BGE-base-v1.5 (unpaired) & C & 60.6\tiny{$\pm$0.5} & 7.8\tiny{$\pm$0.2} & 5.7\tiny{$\pm$0.7} & 16.1\tiny{$\pm$3.2} & 7.8\tiny{$\pm$15.5} & 3.0\tiny{$\pm$1.2} \\
 & PFR & 66.8\tiny{$\pm$0.3} & 8.3\tiny{$\pm$0.3} & 4.2\tiny{$\pm$0.1} & 30.7\tiny{$\pm$3.0} & 17.7\tiny{$\pm$2.0} & 9.3\tiny{$\pm$0.4} \\
\bottomrule
\end{tabular}}
\end{table*}

%% file: tables/per_task.tex
\begin{table*}[!t]
\caption{Per-task chance-normalised accuracy ($\times100$; mean over seeds; question groups averaged within a task).
The training objective is given in parentheses.}\label{tab:pertask}
\centering\small
\setlength{\tabcolsep}{3pt}
\fittable{\begin{tabular}{@{}l cccccccc ccc@{}}
\toprule
 & \multicolumn{8}{c}{Seen (grounded training tasks)} & \multicolumn{3}{c}{Held-out} \\
\cmidrule(lr){2-9}\cmidrule(lr){10-12}
Model & PubMedQA & SciFact & HealthVer & DDI & HoC & ADE & Druglib & MEDLINE & PUBHEALTH & BIOSSES & MTSamples \\
\midrule
PubMedBERT C (PG) & 51.6 & 51.2 & 48.8 & 89.3 & 88.1 & 87.3 & 35.6 & 86.5 & 23.1 & 0.4 & 65.2 \\
S-PubMedBERT C (PG) & 47.6 & 55.3 & 49.1 & 91.0 & 88.7 & 88.0 & 35.4 & 86.4 & 24.2 & 5.4 & 68.7 \\
S-PubMedBERT C (CE) & 47.7 & 60.6 & 55.0 & 92.4 & 89.1 & 88.1 & 41.6 & 87.0 & 20.5 & 1.7 & 70.9 \\
S-PubMedBERT \pfr{} (PG) & 44.9 & 34.9 & 52.9 & 91.5 & 85.8 & 86.9 & 42.8 & 82.7 & 17.0 & 2.5 & 51.3 \\
S-PubMedBERT \pfr{} (CE) & 46.9 & 39.9 & 61.9 & 93.3 & 87.3 & 87.3 & 45.7 & 83.8 & 8.4 & 2.5 & 51.5 \\
MedCPT-Query \pfr{} (PG) & 33.1 & 33.8 & 47.9 & 83.5 & 85.5 & 83.2 & 32.6 & 83.4 & 11.4 & 5.4 & 54.0 \\
all-MPNet \pfr{} (PG) & 39.9 & 40.0 & 65.0 & 93.2 & 86.2 & 85.6 & 47.9 & 84.9 & 25.7 & $-$0.4 & 48.9 \\
BioClinical-ModernBERT C (PG) & 48.5 & 66.0 & 64.6 & 93.9 & 88.9 & 87.7 & 48.6 & 86.6 & 19.6 & 13.7 & 62.1 \\
Laya-large C (PG) & 50.3 & 71.8 & 69.3 & 94.0 & 89.5 & 88.0 & 51.3 & 85.6 & 26.6 & 25.0 & 55.7 \\
\midrule
Qwen3.8-27B (zero-shot) & 56.8 & 77.1 & 37.5 & 17.9 & 82.7 & 60.3 & 50.2 & 76.2 & 53.3 & 10.0 & 75.4 \\
Gemma-4-31B (zero-shot) & 62.8 & 80.4 & 47.5 & 67.4 & 79.4 & 60.0 & 51.0 & 77.2 & 54.3 & 27.5 & 76.0 \\
\bottomrule
\end{tabular}}
\end{table*}

%% file: tables/robustness.tex
\begin{table*}[!t]
\caption{Robustness ($\times100$; mean over 3 seeds where available). Perm.\ flip: share of answers
that change under a random option permutation (5 permutations), averaged over all evaluation splits; on the seen tasks alone C heads flip 2--3\% and \pfr{} below 0.5\% (Table~\ref{tab:objfull}). Trial Qs: MIMIC-III
trial-eligibility questions over notes longer than the 512-token context, decided by truncation,
sliding windows (mean of logits or most confident window) or retrieve-then-decide (RtD). Name gate:
MIMIC-IV windows, standard and common-word-name splits. Both LLMs reach AUROC 100.0 on both
name-gate splits.}\label{tab:robust}
\centering\small
\setlength{\tabcolsep}{3.5pt}
\fittable{\begin{tabular}{@{}lc cccc cccc cc@{}}
\toprule
 & Perm. & \multicolumn{4}{c}{Trial Qs: answer (acc$_{\text{cn}}$)} & \multicolumn{4}{c}{Trial Qs: answerable (acc$_{\text{cn}}$)} & \multicolumn{2}{c}{Name gate AUROC} \\
\cmidrule(lr){3-6}\cmidrule(lr){7-10}\cmidrule(lr){11-12}
Model & flip\,\% & trunc & win-mean & win-conf & RtD & trunc & win-mean & win-conf & RtD & test & common-word \\
\midrule
PubMedBERT C & 16.6 & 2.4 & 3.4 & 2.3 & 5.9 & 4.5 & $-$3.7 & $-$2.8 & 7.9 & 72.9 & 71.8 \\
S-PubMedBERT C & 15.7 & 0.0 & $-$3.8 & 0.1 & $-$1.1 & 7.1 & 1.0 & 4.0 & 20.2 & 72.6 & 71.4 \\
S-PubMedBERT PFR & 0.3 & 8.2 & 15.0 & 16.0 & 7.1 & 4.7 & $-$0.6 & $-$1.5 & 5.8 & 74.8 & 74.5 \\
S-PubMedBERT PFR, frozen & 13.1 & $-$0.5 & $-$0.2 & 2.9 & 0.4 & 7.8 & 0.4 & 2.3 & 29.9 & 75.0 & 74.1 \\
S-PubMedBERT B & 0.0 & 10.9 & 19.7 & 20.5 & 8.3 & 9.8 & 4.2 & 4.7 & 3.8 & 77.6 & 77.7 \\
all-MPNet C & 13.8 & 8.5 & 12.4 & 11.8 & 8.7 & 4.0 & $-$5.9 & $-$5.3 & 5.2 & 72.1 & 71.1 \\
all-MPNet PFR & 0.2 & $-$2.1 & 0.5 & 0.7 & 0.8 & 5.1 & 0.7 & 2.8 & 4.6 & 61.3 & 62.3 \\
S-PubMedBERT PFR, +clinical & 0.3 & 11.1 & 17.8 & 15.8 & 16.0 & $-$1.3 & $-$2.9 & $-$2.9 & 6.9 & 99.8 & 98.5 \\
Laya-large FT & 9.8 & 10.9 & 14.9 & 20.8 & 8.7 & 6.9 & $-$0.8 & 0.6 & 6.0 & 84.9 & 83.4 \\
Laya-large zero-shot & 18.9 & 3.6 & $-$1.8 & $-$0.1 & 0.5 & 5.0 & $-$3.8 & $-$0.6 & $-$2.6 & 84.0 & 83.2 \\
\bottomrule
\end{tabular}}
\end{table*}

%% file: tables/sensitivity.tex
\ifdefined\pairedappendixtables\else\begin{table}[!t]\fi
\caption{Seen-task accuracy sensitivity to removal of exact test inputs appearing in public
training/calibration data, and repeated test inputs. Seed means, in chance-normalised points.
Checkpoints and temperatures are unchanged; this does not repair calibration contamination.}\label{tab:sensitivity}
\centering\small
\fittable{\begin{tabular}{@{}lrrr@{}}\toprule
Model & Historical & Filtered & Difference \\
\midrule
S-PubMedBERT C & 67.69 & 67.32 & $-$0.37 \\
S-PubMedBERT PFR (PG) & 65.30 & 64.91 & $-$0.40 \\
S-PubMedBERT PFR (CE) & 68.26 & 67.89 & $-$0.37 \\
S-PubMedBERT PFR (Reparam) & 68.53 & 68.19 & $-$0.35 \\
BioClinical-ModernBERT C & 73.10 & 72.81 & $-$0.29 \\
Laya-large FT & 74.99 & 74.69 & $-$0.30 \\
Gemma-4-31B & 65.72 & 65.38 & $-$0.34 \\
\bottomrule\end{tabular}}
\ifdefined\pairedappendixtables\else\end{table}\fi

%% file: references.bib
@misc{typesafe2026jev,
  title        = {Introducing System One Models and {Jev}},
  author       = {{TypeSafe AI}},
  year         = {2026},
  howpublished = {\url{https://typesafe.ai/blog/introducing-system-one-models-and-jev}},
  note         = {Accessed 2026-09-29}
}

@misc{laya2026,
  title        = {Laya: A Non-autoregressive System 1 Decision Engine},
  author       = {{Convai Innovations}},
  year         = {2026},
  howpublished = {\url{https://github.com/NandhaKishorM/laya}},
  note         = {Model: \url{https://huggingface.co/convaiinnovations/laya}. Accessed 2026-09-29}
}

@misc{decider2026,
  title        = {decider: One-Pass Typed Decisions with Calibrated Probabilities},
  author       = {{Mapika}},
  year         = {2026},
  howpublished = {\url{https://github.com/Mapika/decider}},
  note         = {Accessed 2026-09-29}
}

@misc{decisionindex2026,
  title        = {{Decision Index}},
  author       = {{apolinario}},
  year         = {2026},
  howpublished = {\url{https://github.com/apolinario/decision-index}},
  note         = {Benchmark suite for typed decision models, edition 0.2.1 (2026-09-27)}
}

@misc{pentest2026,
  howpublished = {arXiv preprint arXiv:2609.28940},
  title         = {Calibrated Decision Models for Autonomous Penetration-Testing Harnesses: {JEV} and {Laya} as System One Decision Layers for {LLM}-Driven Pentest Agents},
  author        = {dos Santos Barbosa, Joas Antonio},
  year          = {2026},
  eprint        = {2609.28940},
  archivePrefix = {arXiv}
}

@misc{css2026,
  howpublished = {arXiv preprint arXiv:2609.24574},
  title         = {Evaluating Decision Models for Text Annotation in Computational Social Science},
  author        = {Ibrahim, Hazem and Zaki, Yasir},
  year          = {2026},
  eprint        = {2609.24574},
  archivePrefix = {arXiv}
}

@misc{crash2026,
  howpublished = {arXiv preprint arXiv:2609.24052},
  title         = {Calibrated Decisions at Scale: Converting Police Crash Narratives into Probabilistic Crash Variables with a System One Model ({Jev})},
  author        = {Rafe, Amir and Das, Subasish},
  year          = {2026},
  eprint        = {2609.24052},
  archivePrefix = {arXiv}
}

@misc{jevout2026,
  howpublished = {arXiv preprint arXiv:2609.30243},
  title         = {{JevOut}: Natural Context Can Flip Decision Models},
  author        = {Xu, Zixiang and Song, Zirui and Zhang, Chiyu and Chen, Xiuying and Liu, Xi and Hu, Xiyang and Zhao, Yue},
  year          = {2026},
  eprint        = {2609.30243},
  archivePrefix = {arXiv}
}

@misc{lavoir2026,
  howpublished = {arXiv preprint arXiv:2609.30706},
  title         = {{LAVOIR}: Teaching a Single-Pass Decision Encoder When and What to Ask with Amortized Value of Information},
  author        = {Yilmaz, Furkan and Tasdemir, Habibe Aleyna and Gozay, Muhammed Faruk},
  year          = {2026},
  eprint        = {2609.30706},
  archivePrefix = {arXiv}
}

@article{gneiting2007strictly,
  title   = {Strictly Proper Scoring Rules, Prediction, and Estimation},
  author  = {Gneiting, Tilmann and Raftery, Adrian E.},
  journal = {Journal of the American Statistical Association},
  volume  = {102},
  number  = {477},
  pages   = {359--378},
  year    = {2007},
  doi = {10.1198/016214506000001437}
}

@article{brier1950,
  title   = {Verification of Forecasts Expressed in Terms of Probability},
  author  = {Brier, Glenn W.},
  journal = {Monthly Weather Review},
  volume  = {78},
  number  = {1},
  pages   = {1--3},
  year    = {1950},
  doi = {10.1175/1520-0493(1950)078<0001:VOFEIT>2.0.CO;2}
}

@article{epstein1969,
  title   = {A Scoring System for Probability Forecasts of Ranked Categories},
  author  = {Epstein, Edward S.},
  journal = {Journal of Applied Meteorology},
  volume  = {8},
  number  = {6},
  pages   = {985--987},
  year    = {1969},
  doi = {10.1175/1520-0450(1969)008<0985:ASSFPF>2.0.CO;2}
}

@inproceedings{guo2017calibration,
  title     = {On Calibration of Modern Neural Networks},
  author    = {Guo, Chuan and Pleiss, Geoff and Sun, Yu and Weinberger, Kilian Q.},
  booktitle = {Proceedings of the 34th International Conference on Machine Learning},
  pages     = {1321--1330},
  year      = {2017},
  url = {https://proceedings.mlr.press/v70/guo17a.html}
}

@inproceedings{desai2020calibration,
  title     = {Calibration of Pre-trained Transformers},
  author    = {Desai, Shrey and Durrett, Greg},
  booktitle = {Proceedings of the 2020 Conference on Empirical Methods in Natural Language Processing},
  pages     = {295--302},
  year      = {2020},
  doi = {10.18653/v1/2020.emnlp-main.21}
}

@article{shao2024deepseekmath,
  title   = {{DeepSeekMath}: Pushing the Limits of Mathematical Reasoning in Open Language Models},
  author  = {Shao, Zhihong and Wang, Peiyi and Zhu, Qihao and Xu, Runxin and Song, Junxiao and Zhang, Mingchuan and Li, Y. K. and Wu, Y. and Guo, Daya},
  journal = {arXiv preprint arXiv:2402.03300},
  year    = {2024},
  url = {https://arxiv.org/abs/2402.03300}
}

@inproceedings{warner2024modernbert,
  title     = {Smarter, Better, Faster, Longer: A Modern Bidirectional Encoder for Fast, Memory Efficient, and Long Context Finetuning and Inference},
  author    = {Warner, Benjamin and Chaffin, Antoine and Clavi{\'e}, Benjamin and Weller, Orion and Hallstr{\"o}m, Oskar and Taghadouini, Said and Gallagher, Alexis and Biswas, Raja and Ladhak, Faisal and Aarsen, Tom and Cooper, Nathan and Adams, Griffin and Howard, Jeremy and Poli, Iacopo},
  booktitle = {Proceedings of the 63rd Annual Meeting of the Association for Computational Linguistics (Volume 1: Long Papers)},
  year      = {2025},
  doi       = {10.18653/v1/2025.acl-long.127}
}

@inproceedings{reimers2019sentence,
  title     = {Sentence-{BERT}: Sentence Embeddings using {S}iamese {BERT}-Networks},
  author    = {Reimers, Nils and Gurevych, Iryna},
  booktitle = {Proceedings of the 2019 Conference on Empirical Methods in Natural Language Processing and the 9th International Joint Conference on Natural Language Processing},
  pages     = {3982--3992},
  year      = {2019},
  doi = {10.18653/v1/D19-1410}
}

@article{nogueira2019passage,
  title   = {Passage Re-ranking with {BERT}},
  author  = {Nogueira, Rodrigo and Cho, Kyunghyun},
  journal = {arXiv preprint arXiv:1901.04085},
  year    = {2019},
  url = {https://arxiv.org/abs/1901.04085}
}

@inproceedings{yin2019benchmarking,
  title     = {Benchmarking Zero-shot Text Classification: Datasets, Evaluation and Entailment Approach},
  author    = {Yin, Wenpeng and Hay, Jamaal and Roth, Dan},
  booktitle = {Proceedings of the 2019 Conference on Empirical Methods in Natural Language Processing and the 9th International Joint Conference on Natural Language Processing},
  pages     = {3914--3923},
  year      = {2019},
  doi = {10.18653/v1/D19-1404}
}

@article{gu2021domain,
  title   = {Domain-Specific Language Model Pretraining for Biomedical Natural Language Processing},
  author  = {Gu, Yu and Tinn, Robert and Cheng, Hao and Lucas, Michael and Usuyama, Naoto and Liu, Xiaodong and Naumann, Tristan and Gao, Jianfeng and Poon, Hoifung},
  journal = {ACM Transactions on Computing for Healthcare},
  volume  = {3},
  number  = {1},
  pages   = {1--23},
  year    = {2021},
  doi = {10.1145/3458754}
}

@article{jin2023medcpt,
  title   = {{MedCPT}: Contrastive Pre-trained Transformers with Large-scale {PubMed} Search Logs for Zero-shot Biomedical Information Retrieval},
  author  = {Jin, Qiao and Kim, Won and Chen, Qingyu and Comeau, Donald C. and Yeganova, Lana and Wilbur, W. John and Lu, Zhiyong},
  journal = {Bioinformatics},
  volume  = {39},
  number  = {11},
  pages   = {btad651},
  year    = {2023},
  doi = {10.1093/bioinformatics/btad651}
}

@article{deka2022improved,
  title   = {Improved Methods To Aid Unsupervised Evidence-Based Fact Checking For Online Health News},
  author  = {Deka, Pritam and Jurek-Loughrey, Anna and Deepak, P.},
  journal = {Journal of Data Intelligence},
  volume  = {3},
  number  = {4},
  pages   = {474--504},
  year    = {2022},
  doi     = {10.26421/jdi3.4-5}
}

@inproceedings{song2020mpnet,
  title     = {{MPNet}: Masked and Permuted Pre-training for Language Understanding},
  author    = {Song, Kaitao and Tan, Xu and Qin, Tao and Lu, Jianfeng and Liu, Tie-Yan},
  booktitle = {Advances in Neural Information Processing Systems},
  volume    = {33},
  year      = {2020},
  url       = {https://proceedings.neurips.cc/paper/2020/hash/c3a690be93aa602ee2dc0ccab5b7b67e-Abstract.html}
}

@inproceedings{jin2019pubmedqa,
  title     = {{PubMedQA}: A Dataset for Biomedical Research Question Answering},
  author    = {Jin, Qiao and Dhingra, Bhuwan and Liu, Zhengping and Cohen, William and Lu, Xinghua},
  booktitle = {Proceedings of the 2019 Conference on Empirical Methods in Natural Language Processing and the 9th International Joint Conference on Natural Language Processing},
  pages     = {2567--2577},
  year      = {2019},
  doi = {10.18653/v1/D19-1259}
}

@article{jin2021disease,
  title   = {What Disease Does This Patient Have? {A} Large-Scale Open Domain Question Answering Dataset from Medical Exams},
  author  = {Jin, Di and Pan, Eileen and Oufattole, Nassim and Weng, Wei-Hung and Fang, Hanyi and Szolovits, Peter},
  journal = {Applied Sciences},
  volume  = {11},
  number  = {14},
  pages   = {6421},
  year    = {2021},
  doi = {10.3390/app11146421}
}

@inproceedings{pal2022medmcqa,
  title     = {{MedMCQA}: A Large-scale Multi-Subject Multi-Choice Dataset for Medical domain Question Answering},
  author    = {Pal, Ankit and Umapathi, Logesh Kumar and Sankarasubbu, Malaikannan},
  booktitle = {Proceedings of the Conference on Health, Inference, and Learning},
  pages     = {248--260},
  year      = {2022},
  url = {https://proceedings.mlr.press/v174/pal22a.html}
}

@inproceedings{hendrycks2021measuring,
  title     = {Measuring Massive Multitask Language Understanding},
  author    = {Hendrycks, Dan and Burns, Collin and Basart, Steven and Zou, Andy and Mazeika, Mantas and Song, Dawn and Steinhardt, Jacob},
  booktitle = {International Conference on Learning Representations},
  year      = {2021},
  url = {https://openreview.net/forum?id=d7KBjmI3GmQ}
}

@inproceedings{wadden2020fact,
  title     = {Fact or Fiction: Verifying Scientific Claims},
  author    = {Wadden, David and Lin, Shanchuan and Lo, Kyle and Wang, Lucy Lu and van Zuylen, Madeleine and Cohan, Arman and Hajishirzi, Hannaneh},
  booktitle = {Proceedings of the 2020 Conference on Empirical Methods in Natural Language Processing},
  pages     = {7534--7550},
  year      = {2020},
  doi = {10.18653/v1/2020.emnlp-main.609}
}

@inproceedings{sarrouti2021evidence,
  title     = {Evidence-based Fact-Checking of Health-related Claims},
  author    = {Sarrouti, Mourad and Ben Abacha, Asma and M'rabet, Yassine and Demner-Fushman, Dina},
  booktitle = {Findings of the Association for Computational Linguistics: EMNLP 2021},
  pages     = {3499--3512},
  year      = {2021},
  doi = {10.18653/v1/2021.findings-emnlp.297}
}

@inproceedings{kotonya2020explainable,
  title     = {Explainable Automated Fact-Checking for Public Health Claims},
  author    = {Kotonya, Neema and Toni, Francesca},
  booktitle = {Proceedings of the 2020 Conference on Empirical Methods in Natural Language Processing},
  pages     = {7740--7754},
  year      = {2020},
  doi = {10.18653/v1/2020.emnlp-main.623}
}

@article{herrero2013ddi,
  title   = {The {DDI} Corpus: An Annotated Corpus with Pharmacological Substances and Drug--Drug Interactions},
  author  = {Herrero-Zazo, Mar{\'\i}a and Segura-Bedmar, Isabel and Mart{\'\i}nez, Paloma and Declerck, Thierry},
  journal = {Journal of Biomedical Informatics},
  volume  = {46},
  number  = {5},
  pages   = {914--920},
  year    = {2013},
  doi = {10.1016/j.jbi.2013.07.011}
}

@article{baker2016automatic,
  title   = {Automatic Semantic Classification of Scientific Literature According to the Hallmarks of Cancer},
  author  = {Baker, Simon and Silins, Ilona and Guo, Yufan and Ali, Imran and H{\"o}gberg, Johan and Stenius, Ulla and Korhonen, Anna},
  journal = {Bioinformatics},
  volume  = {32},
  number  = {3},
  pages   = {432--440},
  year    = {2016},
  doi = {10.1093/bioinformatics/btv585}
}

@article{gurulingappa2012development,
  title   = {Development of a Benchmark Corpus to Support the Automatic Extraction of Drug-related Adverse Effects from Medical Case Reports},
  author  = {Gurulingappa, Harsha and Rajput, Abdul Mateen and Roberts, Angus and Fluck, Juliane and Hofmann-Apitius, Martin and Toldo, Luca},
  journal = {Journal of Biomedical Informatics},
  volume  = {45},
  number  = {5},
  pages   = {885--892},
  year    = {2012},
  doi = {10.1016/j.jbi.2012.04.008}
}

@inproceedings{grasser2018aspect,
  title     = {Aspect-Based Sentiment Analysis of Drug Reviews Applying Cross-Domain and Cross-Data Learning},
  author    = {Gr{\"a}{\ss}er, Felix and Kallumadi, Surya and Malberg, Hagen and Zaunseder, Sebastian},
  booktitle = {Proceedings of the 2018 International Conference on Digital Health},
  pages     = {121--125},
  year      = {2018},
  doi = {10.1145/3194658.3194677}
}

@article{sogancioglu2017biosses,
  title   = {{BIOSSES}: A Semantic Sentence Similarity Estimation System for the Biomedical Domain},
  author  = {So{\u{g}}anc{\i}o{\u{g}}lu, Gizem and {\"O}zt{\"u}rk, Hakime and {\"O}zg{\"u}r, Arzucan},
  journal = {Bioinformatics},
  volume  = {33},
  number  = {14},
  pages   = {i49--i58},
  year    = {2017},
  doi = {10.1093/bioinformatics/btx238}
}

@inproceedings{romanov2018lessons,
  title     = {Lessons from Natural Language Inference in the Clinical Domain},
  author    = {Romanov, Alexey and Shivade, Chaitanya},
  booktitle = {Proceedings of the 2018 Conference on Empirical Methods in Natural Language Processing},
  pages     = {1586--1596},
  year      = {2018},
  doi = {10.18653/v1/D18-1187}
}

@article{johnson2016mimic,
  title   = {{MIMIC-III}, a Freely Accessible Critical Care Database},
  author  = {Johnson, Alistair E. W. and Pollard, Tom J. and Shen, Lu and Lehman, Li-wei H. and Feng, Mengling and Ghassemi, Mohammad and Moody, Benjamin and Szolovits, Peter and Celi, Leo Anthony and Mark, Roger G.},
  journal = {Scientific Data},
  volume  = {3},
  pages   = {160035},
  year    = {2016},
  doi = {10.1038/sdata.2016.35}
}

@article{johnson2023mimiciv,
  title   = {{MIMIC-IV}, a Freely Accessible Electronic Health Record Dataset},
  author  = {Johnson, Alistair E. W. and Bulgarelli, Lucas and Shen, Lu and Gayles, Alvin and Shammout, Ayad and Horng, Steven and Pollard, Tom J. and Hao, Sicheng and Moody, Benjamin and Gow, Brian and Lehman, Li-wei H. and Celi, Leo A. and Mark, Roger G.},
  journal = {Scientific Data},
  volume  = {10},
  pages   = {1},
  year    = {2023},
  doi = {10.1038/s41597-022-01899-x}
}

@article{woo2024synthetic,
  title   = {Synthetic Data Distillation Enables the Extraction of Clinical Information at Scale},
  author  = {Woo, Elizabeth Geena and Burkhart, Michael C. and Alsentzer, Emily and Beaulieu-Jones, Brett K.},
  journal = {npj Digital Medicine},
  volume  = {8},
  pages   = {267},
  doi     = {10.1038/s41746-025-01681-4},
  year    = {2025}
}

@misc{woo2025trialq,
  title        = {{MIMIC-III-Ext-Synthetic-Clinical-Trial-Questions} (version 1.0.0)},
  author       = {Woo, Elizabeth and Burkhart, Michael Craig and Alsentzer, Emily and Beaulieu-Jones, Brett},
  year         = {2025},
  howpublished = {PhysioNet. RRID:SCR\_007345},
  doi          = {10.13026/30k0-av04}
}

@misc{shivade2019mednli,
  title        = {{MedNLI} -- A Natural Language Inference Dataset For The Clinical Domain (version 1.0.0)},
  author       = {Shivade, Chaitanya},
  year         = {2019},
  howpublished = {PhysioNet. RRID:SCR\_007345},
  doi          = {10.13026/C2RS98}
}

@article{pollard2026physionet,
  title   = {{PhysioNet} as a Global Platform for Biomedical Research},
  author  = {Pollard, Tom and Moody, Benjamin E. and Lehman, Li-wei and Gow, Brian and Fernandes, Chrystinne and Xie, Chen and Johnson, Alistair and Mark, Roger G. and Heldt, Thomas},
  journal = {Nature Health},
  year    = {2026},
  doi     = {10.1038/s44360-026-00096-z}
}

@misc{deidnames2023,
  title        = {Annotated {MIMIC-IV} Discharge Summaries for a Study on Deidentification of Names (version 1.0)},
  author       = {Lim, Shulammite and Xiao, Yuxin and Johnson, Alistair and Moukheiber, Dana and Moukheiber, Lama and Moukheiber, Mira and Ghassemi, Marzyeh and Pollard, Tom},
  year         = {2023},
  howpublished = {PhysioNet},
  doi          = {10.13026/63ab-qf77}
}

@inproceedings{thakur2021beir,
  title     = {{BEIR}: A Heterogeneous Benchmark for Zero-shot Evaluation of Information Retrieval Models},
  author    = {Thakur, Nandan and Reimers, Nils and R{\"u}ckl{\'e}, Andreas and Srivastava, Abhishek and Gurevych, Iryna},
  booktitle = {Thirty-fifth Conference on Neural Information Processing Systems Datasets and Benchmarks Track},
  year      = {2021}
}

@inproceedings{boteva2016full,
  title     = {A Full-Text Learning to Rank Dataset for Medical Information Retrieval},
  author    = {Boteva, Vera and Gholipour, Demian and Sokolov, Artem and Riezler, Stefan},
  booktitle = {Advances in Information Retrieval (ECIR 2016)},
  pages     = {716--722},
  year      = {2016},
  doi = {10.1007/978-3-319-30671-1_58}
}

@article{chen2023frugalgpt,
  title   = {{FrugalGPT}: How to Use Large Language Models While Reducing Cost and Improving Performance},
  author  = {Chen, Lingjiao and Zaharia, Matei and Zou, James},
  journal = {Transactions on Machine Learning Research},
  year    = {2024},
  url     = {https://openreview.net/forum?id=cSimKw5p6R}
}

@article{nussbaum2024nomic,
  title   = {{Nomic Embed}: Training a Reproducible Long Context Text Embedder},
  author  = {Nussbaum, Zach and Morris, John Xavier and Mulyar, Andriy and Duderstadt, Brandon},
  journal = {Transactions on Machine Learning Research},
  year    = {2025},
  url     = {https://openreview.net/forum?id=IPmzyQSiQE}
}

@inproceedings{zhang2024mgte,
  title     = {{mGTE}: Generalized Long-Context Text Representation and Reranking Models for Multilingual Text Retrieval},
  author    = {Zhang, Xin and Zhang, Yanzhao and Long, Dingkun and Xie, Wen and Dai, Ziqi and Tang, Jialong and Lin, Huan and Yang, Baosong and Xie, Pengjun and Huang, Fei and Zhang, Meishan and Li, Wenjie and Zhang, Min},
  booktitle = {Proceedings of the 2024 Conference on Empirical Methods in Natural Language Processing: Industry Track},
  pages     = {1393--1412},
  year      = {2024},
  doi       = {10.18653/v1/2024.emnlp-industry.103}
}

@article{li2023towards,
 title={Towards general text embeddings with multi-stage contrastive learning},
 author={Li, Zehan and Zhang, Xin and Zhang, Yanzhao and Long, Dingkun and Xie, Pengjun and Zhang, Meishan},
 journal={arXiv preprint arXiv:2308.03281},
 year={2023}
}

@misc{bcmbembed,
 author={{NeuML}},
 title={BioClinical ModernBERT embeddings model card},
 howpublished={\url{https://huggingface.co/NeuML/bioclinical-modernbert-base-embeddings}},
 note={Model identifier used in the experiment configuration},
 year={2025}
}

@misc{sounack2025bioclinicalmodernbertstateoftheartlongcontext,
  howpublished = {arXiv preprint arXiv:2506.10896},
 title={{BioClinical ModernBERT}: A State-of-the-Art Long-Context Encoder for Biomedical and Clinical {NLP}},
 author={Sounack, Thomas and Davis, Joshua and Durieux, Brigitte and Chaffin, Antoine and Pollard, Tom J. and Lehman, Eric and Johnson, Alistair E. W. and McDermott, Matthew and Naumann, Tristan and Lindvall, Charlotta},
 year={2025},
 eprint={2506.10896},
 archivePrefix={arXiv},
 primaryClass={cs.CL},
 url={https://arxiv.org/abs/2506.10896}
}

@misc{allmpnet,
 author={{Sentence Transformers}},
 title={all-MPNet-base-v2 model card},
 howpublished={\url{https://huggingface.co/sentence-transformers/all-mpnet-base-v2}},
 note={Model identifier used in the experiment configuration},
 year={2021}
}

@inproceedings{xiao2024c,
  title     = {{C-Pack}: Packed Resources For General {Chinese} Embeddings},
  author    = {Xiao, Shitao and Liu, Zheng and Zhang, Peitian and Muennighoff, Niklas and Lian, Defu and Nie, Jian-Yun},
  booktitle = {Proceedings of the 47th International ACM SIGIR Conference on Research and Development in Information Retrieval},
  pages     = {641--649},
  year      = {2024},
  doi       = {10.1145/3626772.3657878}
}

@misc{qwen38,
  title  = {{Qwen3.8-Max}: A New Bar for Coding and Cowork},
  author = {{Qwen Team}},
  year   = {2026},
  month  = {August},
  url    = {https://qwen.ai/blog?id=qwen3.8},
  note   = {Citation recommended on the model card of Qwen3.8-27B, \url{https://huggingface.co/Qwen/Qwen3.8-27B}}
}

@misc{gemmateam2026gemma4,
  howpublished = {arXiv preprint arXiv:2607.02770},
 title={{Gemma 4} Technical Report},
 author={{Gemma Team}},
 year={2026},
 eprint={2607.02770},
 archivePrefix={arXiv},
 primaryClass={cs.CL},
 url={https://arxiv.org/abs/2607.02770}
}

@article{lee2020biobert,
  title={{BioBERT}: a pre-trained biomedical language representation model for biomedical text mining},
  author={Lee, Jinhyuk and Yoon, Wonjin and Kim, Sungdong and Kim, Donghyeon and Kim, Sunkyu and So, Chan Ho and Kang, Jaewoo},
  journal={Bioinformatics},
  volume={36},
  number={4},
  pages={1234--1240},
  year={2020},
  publisher={Oxford University Press},
  doi={10.1093/bioinformatics/btz682}
}

@inproceedings{nguyen2016ms,
  title={{MS MARCO}: A Human Generated Machine Reading Comprehension Dataset},
  author={Nguyen, Tri and Rosenberg, Mir and Song, Xia and Gao, Jianfeng and Tiwary, Saurabh and Majumder, Rangan and Deng, Li},
  booktitle={Proceedings of the Workshop on Cognitive Computation: Integrating neural and symbolic approaches 2016 co-located with the 30th Annual Conference on Neural Information Processing Systems (NIPS 2016)},
  series={CEUR Workshop Proceedings},
  volume={1773},
  year={2016},
  url={https://ceur-ws.org/Vol-1773/CoCoNIPS_2016_paper9.pdf}
}

@inproceedings{karpukhin2020dense,
  title     = {Dense Passage Retrieval for Open-Domain Question Answering},
  author    = {Karpukhin, Vladimir and Oguz, Barlas and Min, Sewon and Lewis, Patrick and Wu, Ledell and Edunov, Sergey and Chen, Danqi and Yih, Wen-tau},
  booktitle = {Proceedings of the 2020 Conference on Empirical Methods in Natural Language Processing (EMNLP)},
  pages     = {6769--6781},
  year      = {2020},
  doi       = {10.18653/v1/2020.emnlp-main.550}
}

@inproceedings{khattab2020colbert,
  title     = {{ColBERT}: Efficient and Effective Passage Search via Contextualized Late Interaction over {BERT}},
  author    = {Khattab, Omar and Zaharia, Matei},
  booktitle = {Proceedings of the 43rd International ACM SIGIR Conference on Research and Development in Information Retrieval},
  pages     = {39--48},
  year      = {2020},
  doi       = {10.1145/3397271.3401075}
}

@inproceedings{schick2021exploiting,
  title     = {Exploiting Cloze-Questions for Few-Shot Text Classification and Natural Language Inference},
  author    = {Schick, Timo and Sch{\"u}tze, Hinrich},
  booktitle = {Proceedings of the 16th Conference of the European Chapter of the Association for Computational Linguistics: Main Volume},
  pages     = {255--269},
  year      = {2021},
  doi       = {10.18653/v1/2021.eacl-main.20}
}

@inproceedings{gao2021making,
  title     = {Making Pre-trained Language Models Better Few-shot Learners},
  author    = {Gao, Tianyu and Fisch, Adam and Chen, Danqi},
  booktitle = {Proceedings of the 59th Annual Meeting of the Association for Computational Linguistics and the 11th International Joint Conference on Natural Language Processing (Volume 1: Long Papers)},
  pages     = {3816--3830},
  year      = {2021},
  doi       = {10.18653/v1/2021.acl-long.295}
}

@inproceedings{khashabi2020unifiedqa,
  title     = {{UnifiedQA}: Crossing Format Boundaries with a Single {QA} System},
  author    = {Khashabi, Daniel and Min, Sewon and Khot, Tushar and Sabharwal, Ashish and Tafjord, Oyvind and Clark, Peter and Hajishirzi, Hannaneh},
  booktitle = {Findings of the Association for Computational Linguistics: EMNLP 2020},
  pages     = {1896--1907},
  year      = {2020},
  doi       = {10.18653/v1/2020.findings-emnlp.171}
}

@inproceedings{niculescu2005predicting,
  title     = {Predicting Good Probabilities with Supervised Learning},
  author    = {Niculescu-Mizil, Alexandru and Caruana, Rich},
  booktitle = {Proceedings of the 22nd International Conference on Machine Learning},
  pages     = {625--632},
  year      = {2005},
  doi       = {10.1145/1102351.1102430}
}

@article{williams1992simple,
  title   = {Simple Statistical Gradient-Following Algorithms for Connectionist Reinforcement Learning},
  author  = {Williams, Ronald J.},
  journal = {Machine Learning},
  volume  = {8},
  pages   = {229--256},
  year    = {1992},
  doi     = {10.1007/BF00992696}
}

@inproceedings{ahmadian2024back,
  title     = {Back to Basics: Revisiting {REINFORCE}-Style Optimization for Learning from Human Feedback in {LLMs}},
  author    = {Ahmadian, Arash and Cremer, Chris and Gall{\'e}, Matthias and Fadaee, Marzieh and Kreutzer, Julia and Pietquin, Olivier and {\"U}st{\"u}n, Ahmet and Hooker, Sara},
  booktitle = {Proceedings of the 62nd Annual Meeting of the Association for Computational Linguistics (Volume 1: Long Papers)},
  pages     = {12248--12267},
  year      = {2024},
  doi       = {10.18653/v1/2024.acl-long.662}
}

@misc{dinzinger2026notbetter,
  howpublished  = {arXiv preprint arXiv:2610.32318},
  title         = {System One Models Are Not Better Rerankers},
  author        = {Dinzinger, Michael and Ghosh Dastidar, Kanishka and Granitzer, Michael},
  year          = {2026},
  eprint        = {2610.32318},
  archivePrefix = {arXiv}
}

@misc{wan2026retrieverfree,
  howpublished  = {arXiv preprint arXiv:2610.35868},
  title         = {Retriever-Free Retrieval: Navigating Documents with Decision Models},
  author        = {Wan, Chao and J{\"u}nger, Paul and Du, Beixi and Liu, Bodong and Krishnan, Navaneetha and Weinberger, Kilian Q.},
  year          = {2026},
  eprint        = {2610.35868},
  archivePrefix = {arXiv}
}
